\documentclass[10pt,twocolumn,letterpaper]{article}

\usepackage[pagenumbers]{cvpr} 

\usepackage{colortbl} 
\usepackage{pifont}

\usepackage{algorithm}
\usepackage{algorithmic}
\usepackage{xcolor}   
\usepackage{tabularx}

\newcommand{\cmark}{\textcolor{green!70!black}{\ding{51}}}
\newcommand{\xmark}{\textcolor{red!80!black}{\ding{55}}}
\usepackage{multirow}
\usepackage{makecell}

\definecolor{cvprblue}{rgb}{0.21,0.49,0.74}
\usepackage[pagebackref,breaklinks,colorlinks,allcolors=cvprblue]{hyperref}
\usepackage{graphicx}
\usepackage{amssymb}
\usepackage{subcaption}

\def\paperID{32372} 
\def\confName{CVPR}
\def\confYear{2026}

\title{E-comIQ-ZH: A Human-Aligned Dataset and Benchmark for Fine-Grained Evaluation of E-commerce Posters with Chain-of-Thought}

\author{
\textbf{Meiqi Sun, Mingyu Li\thanks{Corresponding author: \texttt{lmy194543@taobao.com}}, Junxiong Zhu} \\
Taobao \& Tmall Group, Alibaba Group
}

\begin{document}
\maketitle
\begin{abstract}
Generative AI is widely used to create commercial posters. However, rapid advances in generation have outpaced automated quality assessment. Existing models emphasize generic esthetics or low level distortions and lack the functional criteria required for e-commerce design. It is especially challenging for Chinese content, where complex characters often produce subtle but critical textual artifacts that are overlooked by existing methods. To address this, we introduce E-comIQ-ZH, a framework for evaluating Chinese e-commerce posters. We build the first dataset \textbf{E-comIQ-18k} to feature multi dimensional scores and expert calibrated Chain of Thought (CoT) rationales. Using this dataset, we train \textbf{E-comIQ-M}, a specialized evaluation model that aligns with human expert judgment. Our framework enables \textbf{E-comIQ-Bench}, the first automated and scalable benchmark for the generation of Chinese e-commerce posters. Extensive experiments show our E-comIQ-M aligns more closely with expert standards and enables scalable automated assessment of e-commerce posters. All datasets, models, and evaluation tools will be released to support future research in this area.Code will be available at \url{https://github.com/4mm7/E-comIQ-ZH}.
\end{abstract}

\section{Introduction}
\label{sec:intro}
\begin{figure}[t]
    \centering
    \includegraphics[width=\linewidth]{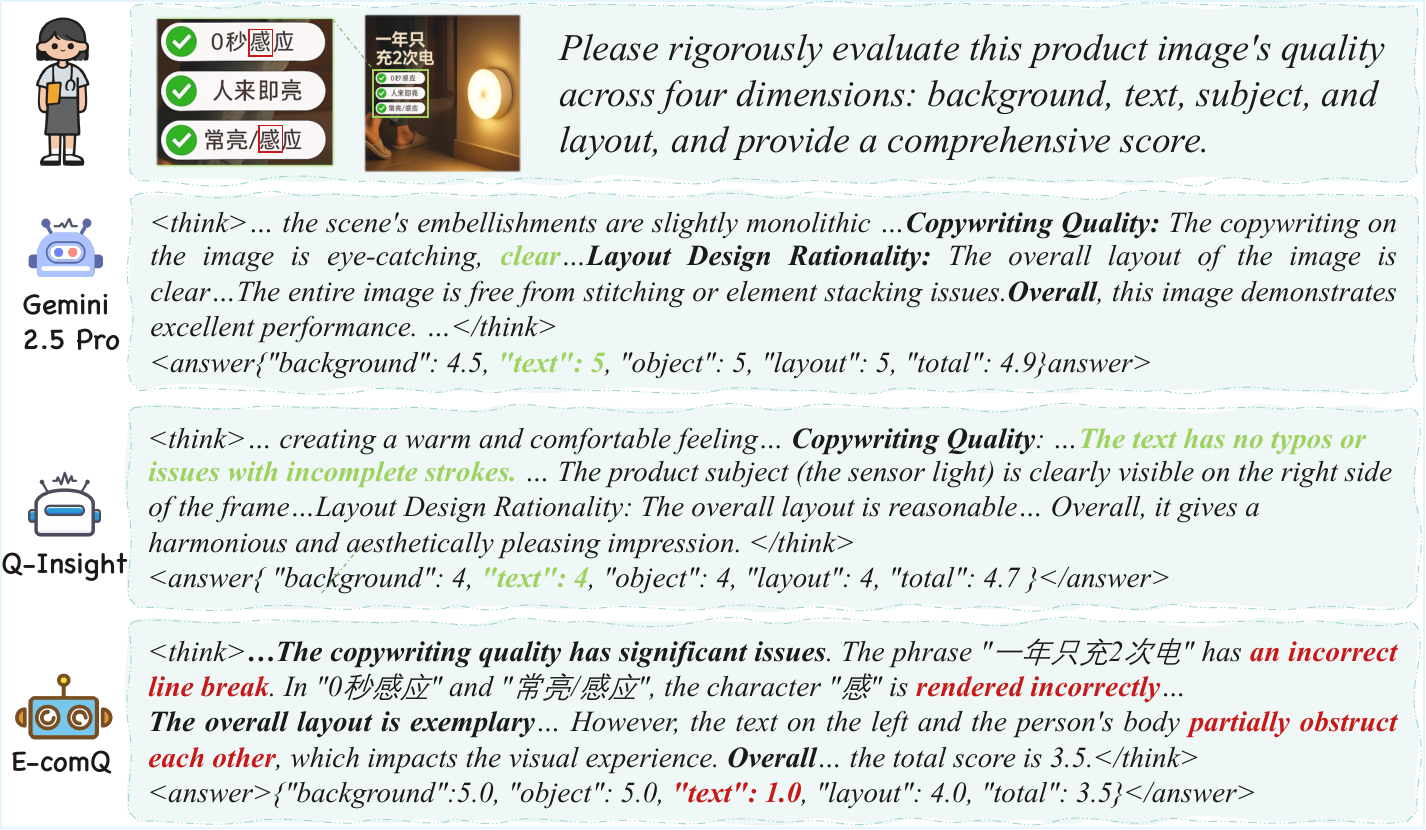}
    \caption{
    \textbf{Qualitative comparison of E-comIQ-M with leading MLLMs on a challenging e-commerce image.} 
        While other powerful models like Gemini 2.5 Pro~\cite{comanici2025gemini} and Q-Insight~\cite{li2025q} overlook critical flaws, our E-comIQ-M accurately identifies the subtle  \textit{stroke-level corruption}.
        This leads to a more human-aligned low score for the text dimension (1.0), demonstrating its superior fine-grained diagnostic capabilities.
    }
    \label{fig:intro}
\end{figure}
Recent advances in generative AI are transforming content creation, with substantial impact on commercial applications~\cite{openai2024sora, seedream2025seedream, batifol2025flux, comanici2025gemini, wu2025qwen}. E-commerce posters are a central case where visuals must blend aesthetic appeal with functional effectiveness~\cite{tian2025ai, xu2025application, wang2025mv, chen2025postercraft}. Recent research shows that large generative models can produce visually appealing posters~\cite{gao2025postermaker, hu2025dreamposter, chen2025t}. However, achieving commercially viable quality with generative models often requires labor-intensive human oversight, including meticulous prompt engineering and iterative refinement~\cite{zhang2025aiguard,zhang2025quality, zhu2025vquala}. This points to a fundamental bottleneck: the lack of automated, reliable Image Quality Assessment (IQA) tools to standardize quality control and guide model optimization~\cite{chahine2024deep, wang2006modern, liu2016blind}. As shown in Fig.~\ref{fig:intro}, prominent general-purpose models~\cite{comanici2025gemini, bai2025qwen2} and emerging IQA methods~\cite{li2025q} tend to focus on general aesthetics, overlooking critical domain-specific flaws. This issue is particularly severe for Chinese e-commerce content. Dense typography and complex characters cause subtle but important text rendering errors.

\begin{figure*}[t]
    \centering
    \includegraphics[width=\linewidth, height=0.28\textheight,keepaspectratio]{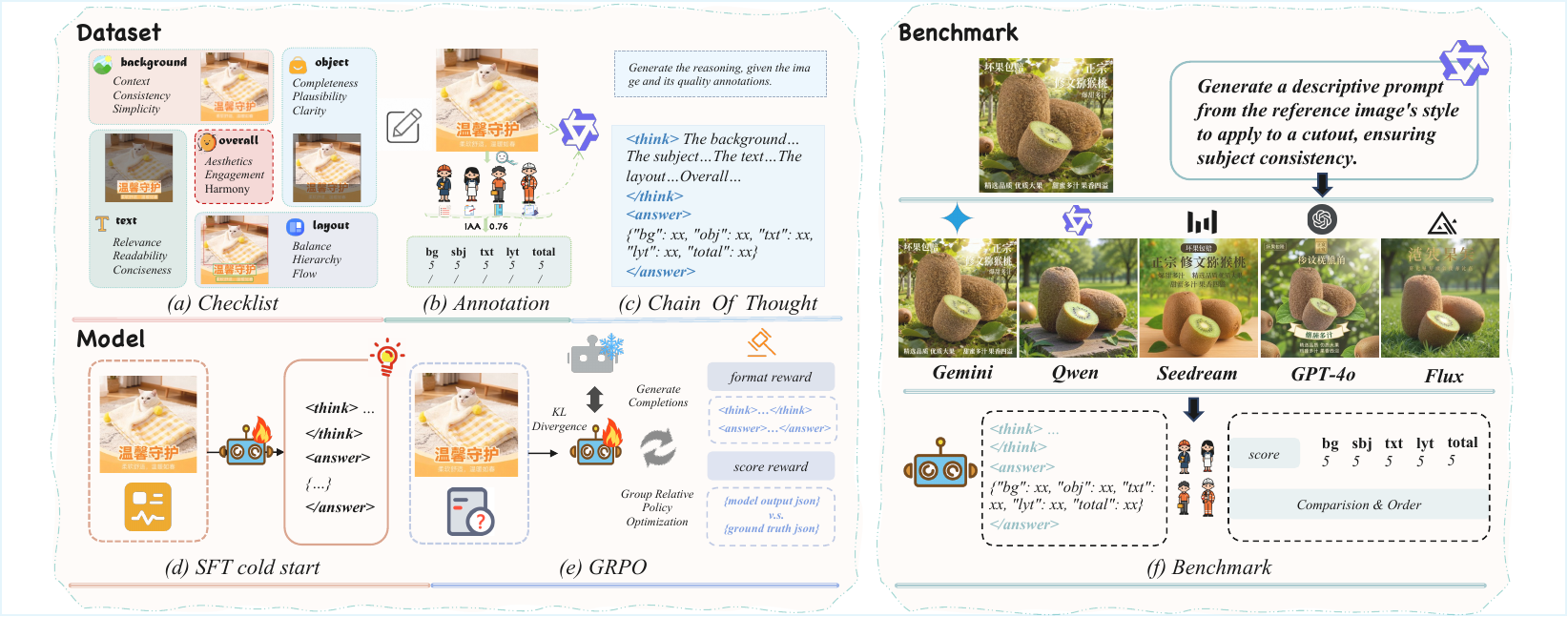}
    \caption{
    \textbf{Overview of the E-comIQ-ZH framework.}
    (a–c) \textbf{E-comIQ-Dataset}: multi-dimensional expert annotations with Chain-of-Thought rationales. 
    (d–e) \textbf{E-comIQ-M}: two-stage training via Supervised Fine-Tuning (SFT) and Generative Reranking Policy Optimization (GRPO). 
    (f) \textbf{E-comIQ-Bench}: evaluation of generative models on e-commerce image generation capabilities.
    }
    \label{fig:overview}
\end{figure*}
Traditional IQA methods mainly target low-level distortions (blur, noise, or compression)~\cite{wang2004image, mittal2012no, bianco2018use}. They cannot judge layout, product visibility, or textual clarity. Recent work on AI-generated e-commerce image quality goes beyond synthetic distortions~\cite{zhang2025aiguard, zhang2025quality, zhu2025vquala}. Several datasets target product quality, background inpainting, or layout~\cite{liang2025evaluation, hsu2023posterlayout, teng2025postercot, wang2024scipostlayout}. However, these efforts typically handle single-product photos or geometric layout, and provide one-dimensional scores or a few defect categories. Multimodal large language models (MLLMs) provide a new class of evaluators. They can perform pairwise comparison~\cite{zhu20242afc, chen2025toward} and holistic scoring~\cite{you2025teaching, wu2023q, li2025q, lu2025q}, and can be further aligned using preference datasets and reward models~\cite{kirstain2023pick, xu2023imagereward}. However, they cannot capture domain-specific questions that matter for e-commerce, such as whether the Chinese copy is accurate and stylistically appropriate.

This misalignment creates a vicious cycle. Without a formal, multi-dimensional quality standard for e-commerce visuals, it is hard to evaluate systems systematically or to build datasets for training specialized evaluators. As a result, current workflows still lack robust automated tools aligned with expert judgment and rely on slow, unscalable manual review. Existing poster generators are often assessed only by internal business metrics or small user studies~\cite{yu2022partiprompts, huang2023t2i}. To address these gaps, we introduce \textbf{E-comIQ-ZH}, illustrated in Fig.~\ref{fig:overview}, with the following contributions:

\begin{itemize}
\item We present \textbf{E-comIQ-18k}, to our knowledge the first large-scale dataset explicitly targeting Chinese e-commerce poster assessment, containing 18{,}000 posters with multi-dimensional functional scores and expert Chain-of-Thought (CoT) rationales.

\item We develop \textbf{E-comIQ-M}, a domain-specific evaluation model aligned with expert judgments and fine-grained e-commerce design criteria, which outperforms general-purpose evaluators on our dataset.

\item We release \textbf{E-comIQ-Bench}, a benchmark for Chinese e-commerce poster generation that enables rigorous and scalable comparison of leading generative models.
\end{itemize}



\vspace{-4pt}
\section{Related Works}
\begin{figure*}[t]
    \centering
    \includegraphics[width=\linewidth]{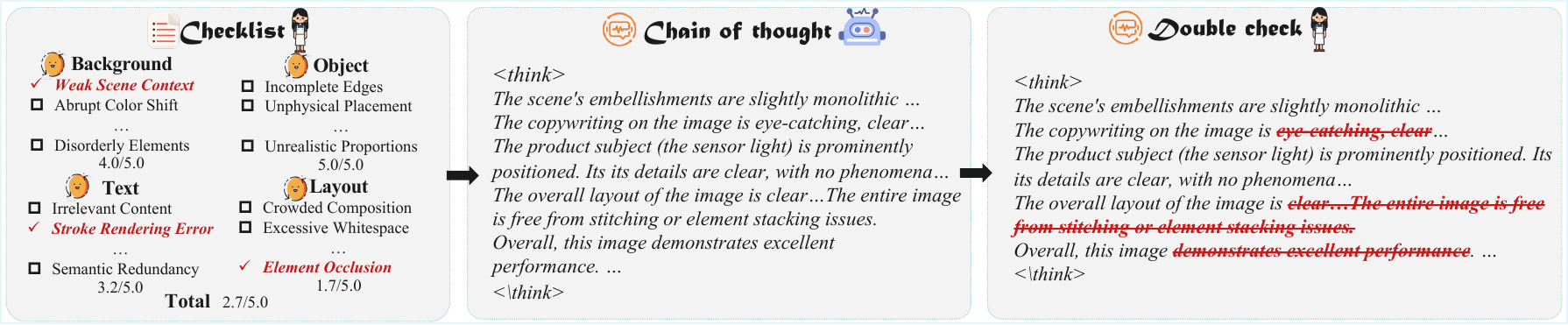}
    \caption{
        \textbf{An illustration of our human-AI collaborative pipeline for generating diagnostic Chain-of-Thought (CoT) rationales.} 
    }
    \label{fig:data_annotation}
\end{figure*}
\vspace{-1pt}
\paragraph{Traditional IQA and Aesthetic Assessment.}
Traditional IQA methods quantify signal fidelity. Full-reference approaches such as SSIM~\cite{wang2004image} measure deviation from a pristine source, while no-reference methods predict quality from handcrafted statistics~\cite{mittal2012no} or deep features~\cite{ke2021musiq, su2020blindly}. Image aesthetic assessment (IAA) instead predicts subjective beauty using large datasets like AVA~\cite{murray2012ava} or LAION-Aesthetics~\cite{schuhmann2022laion}. Both IQA and IAA mainly focus on low-level distortions or generic aesthetics and do not cover the functional, multi-dimensional criteria that determine whether an e-commerce poster is commercially usable.

\vspace{-12pt}
\paragraph{MLLM-based Quality Assessment.}
MLLMs are increasingly used as visual evaluators. Early work fine-tunes MLLMs to output scalar scores~\cite{you2025teaching, wu2024q} or natural-language critiques~\cite{you2024depicting, zhang2024q}. More recent approaches apply preference-based optimization, such as DPO~\cite{rafailov2023direct} and GRPO~\cite{li2025q, wu2025visualquality}. However, they are mostly trained on open-domain data and miss domain-specific criteria such as correctness and readability in e-commerce posters.

\vspace{-12pt}
\paragraph{General IQA Datasets and Benchmarks.}
Benchmark datasets have driven progress in visual assessment, from classical IQA collections with synthetic distortions~\cite{sheikh2006statistical} to large-scale ``in-the-wild'' datasets such as KonIQ-10k and SPAQ~\cite{hosu2020koniq, fang2020perceptual}, and to AIGC benchmarks and preference datasets with learned reward models~\cite{huang2023t2i, kirstain2023pick, xu2023imagereward, xu2024visionreward}. These resources are crucial for training general evaluators, but they mainly provide holistic or binary feedback and lack explanation-rich labels tailored to the e-commerce setting.

\vspace{-12pt}
\paragraph{E-commerce Datasets and Benchmarks.}

Several works address e-commerce visual quality and poster design, proposing product-image assessors and defect detectors~\cite{zhang2025aiguard, zhang2025quality, zhu2025vquala} and datasets or systems for product quality, background editing, and poster/layout design~\cite{liang2025evaluation, hsu2023posterlayout, teng2025postercot, wang2024scipostlayout, gao2025postermaker, chen2025postercraft, hu2025dreamposter}. However, they mainly cover single-product photos or geometric layout with one-dimensional or sparse defect labels, and thus cannot support multi-dimensional evaluation of Chinese e-commerce posters.

\section{E-comIQ-18k}
\begin{figure}[b]
    \centering
    \includegraphics[width=0.3\textwidth]{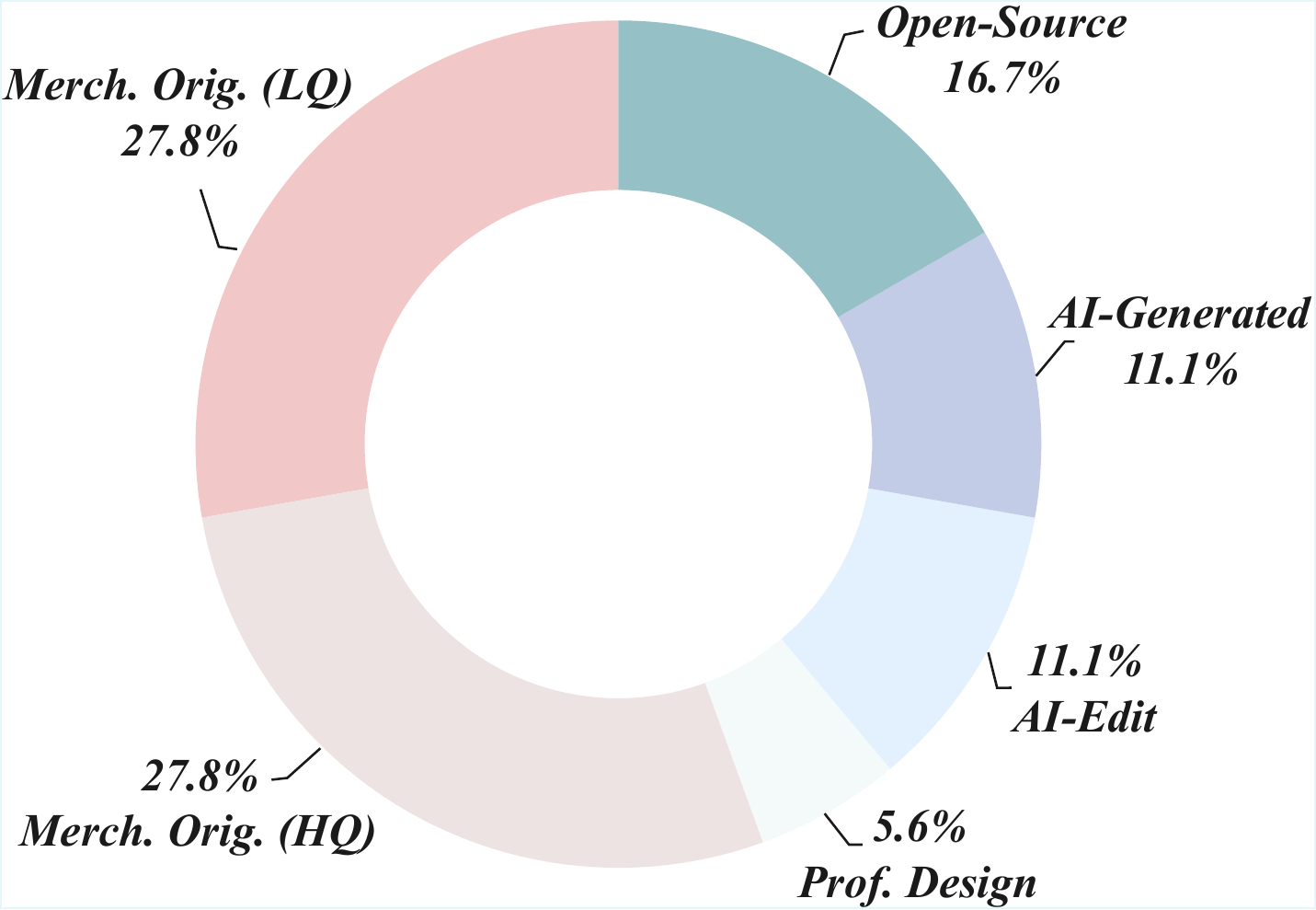}
    \caption{Distribution of image sources.}
    \label{fig:data_source}
\end{figure}

\label{sec:dataset}
\subsection{Dataset Composition and Sourcing}
\label{subsec:collection}

Our E-comIQ-18k dataset comprises 18k images from six distinct sources (see Figure~\ref{fig:data_source}). We first perform a coarse, binary manual screening on a large volume of merchant-provided photos and select 5k high-quality (HQ) and 5k low-quality (LQ) images, capturing a wide spectrum of real-world quality. To establish an upper bound, we include professionally designed images. The dataset is further enriched with two types of synthetic data: AI generated posters created from product cutouts and AI edited compositions that simulate template-based workflows. We split the 18k images into 15k/2k/1k train/validation/test samples with balanced source and quality distributions.

\subsection{Multi-Dimensional Annotation Pipeline}
\label{subsec:annotation}

In collaboration with a panel of senior e-commerce art directors, we decompose e-commerce visual quality into four dimensions. Each image is annotated by a single expert along these dimensions:

\begin{itemize}
    \itemsep0em
    \item Object: visual integrity of the product, including clarity, completeness, and absence of distortion.
    \item Background: compatibility and visual appeal of the background relative to the subject.
    \item Text: legibility and correctness of all typographic elements, as well as their visual integration.
    \item Layout: overall composition, including visual hierarchy and spatial arrangement.
\end{itemize}
For each dimension, annotators provide a continuous score anchored to three quality tiers: excellent [4.0, 5.0], good [3.0, 4.0), and poor [1.0, 3.0). They also select issue tags from a detailed checklist for each dimension.  The complete checklist is provided in the Appendix.

\vspace{-10pt}
\paragraph{Expert Annotation and Quality Assurance.}

The annotation process proceeds in two phases. First, six domain experts independently annotate a shared calibration set of 1{,}000 images. This set is fully cross-reviewed, and disagreements are resolved in consensus meetings until a stable Krippendorff’s Alpha is achieved; as shown in Table~\ref{tab:iaa}, the final overall reliability reaches $\alpha = 0.858$ for both scores and tags. After this calibration, the remaining 17{,}000 images are partitioned among the experts without overlap. To avoid standard drift, we maintain a 10\% random sampling protocol with a shared log for ambiguous cases; additional reliability statistics are reported in the Appendix. An illustration of the human and AI collaborative annotation pipeline is shown in Fig.~\ref{fig:data_annotation}.
 
\begin{table}[b!]
\centering
\small
\caption{
     Inter-annotator agreement for E-comIQ-18k, measured by Krippendorff's Alpha ($\alpha$) and loose accuracy (Acc., within a 0.5 margin), confirming the substantial reliability of our annotations.
}
\label{tab:iaa}
\setlength{\tabcolsep}{4pt} 
\begin{tabular}{lccccc}
\toprule
 & \textbf{Overall} & \textbf{Object} & \textbf{Background} & \textbf{Text} & \textbf{Layout} \\
\midrule
\textbf{$\alpha$} & \textbf{0.858} & 0.745 & 0.721 & 0.765 & 0.877 \\
\textbf{Acc. (\%)} & \textbf{96.4}  & 92.2  & 94.6  & 93.2  & 96.6  \\
\bottomrule
\end{tabular}
\end{table}

\vspace{-10pt}
\paragraph{CoT Generation and Expert Editing.}
To obtain diagnostic CoT rationales at scale, we adopt a human–AI collaborative pipeline. Given the expert scores, issue tags, and the image, we prompt Qwen-2.5-VL-Max~\cite{bai2025qwen2} to generate a rationale that explains the scores and grounds them in concrete visual evidence (prompt template in the Appendix). Each AI-generated rationale is then returned to the original annotator, who uses a NER based interface to delete hallucinated content, correct reasoning errors, and add domain-specific insights. This human supervised process process keeps the scalability of LLM generation while ensuring that the final CoT rationales remain faithful to expert judgments.

\begin{table*}[t!]
\centering
\small 
\caption{
    \textbf{Comparison of E-comIQ-18k with representative image quality, preference, and e-commerce evaluation datasets.}
    Most existing datasets target general aesthetics, distortion fidelity, or holistic AIGC preference, while AIGuard is the only e-commerce functional dataset but relies on binary labels without multidimensional scoring or CoT explanations. E-comIQ-18k uniquely provides e-commerce focused functional multidimensional scores together with expert verified CoT rationales.
}

\label{tab:dataset_comparison}
\begin{tabularx}{\textwidth}{X l l c c c c l} 
\toprule
\textbf{Dataset} & \textbf{Domain} & \textbf{Purpose} & \textbf{\# Images} & \textbf{Ref.} & \textbf{Annotation Type} & \textbf{Score Range} & \textbf{CoT} \\
\midrule

\rowcolor{gray!10}
\multicolumn{8}{l}{\textit{AIGC Quality / Preference}} \\
ImageRewardDB \cite{xu2023imagereward} & AIGC & Preference & ~137k pairs & NR & Pairwise Pref. & N/A & \xmark \\
AGIQA-3K \cite{li2023agiqa} & AIGC & General & 3,000 & NR & Multi-dim. Score & [1, 5] & \xmark \\

\midrule
\rowcolor{gray!10}
\multicolumn{8}{l}{\textit{General ``In-the-Wild''}} \\
KonIQ-10k \cite{hosu2020koniq} & General & General & 10,073 & NR & Single MOS & [1, 5] & \xmark \\
SPAQ \cite{fang2020perceptual} & General & Aesthetics & 11,125 & NR & Single MOS & [1, 100] & \xmark \\
LIVE-FB \cite{ying2020patches} & General & General & ~40,000 & NR & Single MOS & [1, 100] & \xmark \\

\midrule
\rowcolor{gray!10}
\multicolumn{8}{l}{\textit{Synthetic / Full-Reference}} \\
KADID-10k \cite{lin2019kadid} & Synthetic & Fidelity & 10,125 & FR & Single MOS & [1, 5] & \xmark \\
PIPAL \cite{gu2020pipa} & Synthetic & Fidelity & ~29,000 & FR & Single MOS & [1, 5] & \xmark \\

\midrule
\rowcolor{gray!10}
\multicolumn{8}{l}{\textit{E-commerce Visual Quality}} \\
AIGuard~\cite{zhang2025aiguard} & E-commerce & Functional & 253,420 & NR & Binary label+tag  & N/A. & \xmark \\
\textbf{E-comIQ-18k (Ours)} & E-commerce & \textbf{Functional} & \textbf{18,000} & NR & \textbf{Multi-dim. Score} & \textbf{[1, 5]} & \cmark \\

\bottomrule
\end{tabularx}
\end{table*}

\subsection{Dataset Statistics and Properties}
\label{subsec:analysis}
As shown in Table~\ref{tab:dataset_comparison}, most existing image quality datasets focus on general aesthetics or low level fidelity, and the only e-commerce functional dataset AIGuard provides binary labels without multidimensional scoring or explanations. E-comIQ-18k is, to our knowledge, the first large scale dataset that combines an e-commerce focus with functional multidimensional scores and expert verified CoT rationales.

\begin{figure}[t!] 
\centering

\begin{subfigure}[b]{0.23\textwidth}
    \centering
    \includegraphics[width=\linewidth]{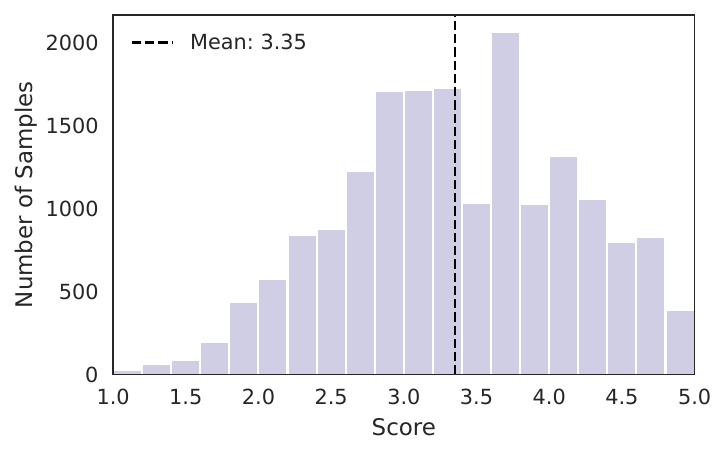} 
    \subcaption{Score Distribution}
    \label{fig:stats_dist}
\end{subfigure}
\hfill 
\begin{subfigure}[b]{0.23\textwidth}
    \centering
    \includegraphics[width=\linewidth]{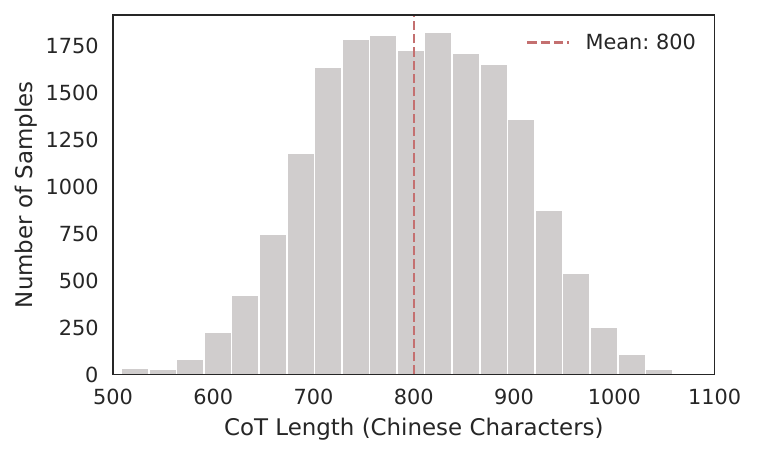} 
    \subcaption{CoT Rationale Length}
    \label{fig:stats_cot}
\end{subfigure}

\vspace{1em} 

\begin{subfigure}[b]{0.23\textwidth}
    \centering
    \includegraphics[width=\linewidth]{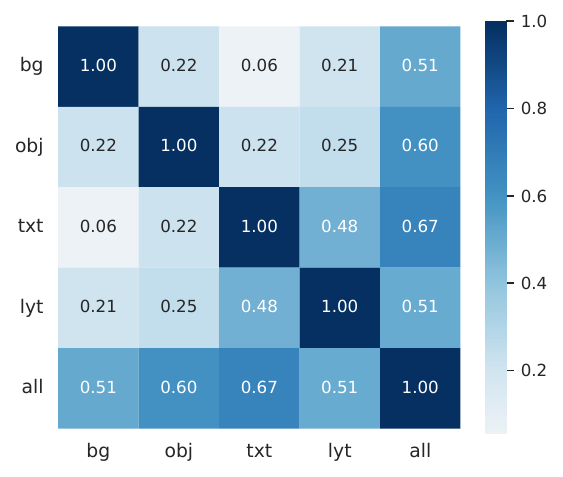} 
    \subcaption{Correlation Matrix}
    \label{fig:stats_corr}
\end{subfigure}
\hfill 
\begin{subfigure}[b]{0.23\textwidth}
    \centering
    \includegraphics[width=\linewidth]{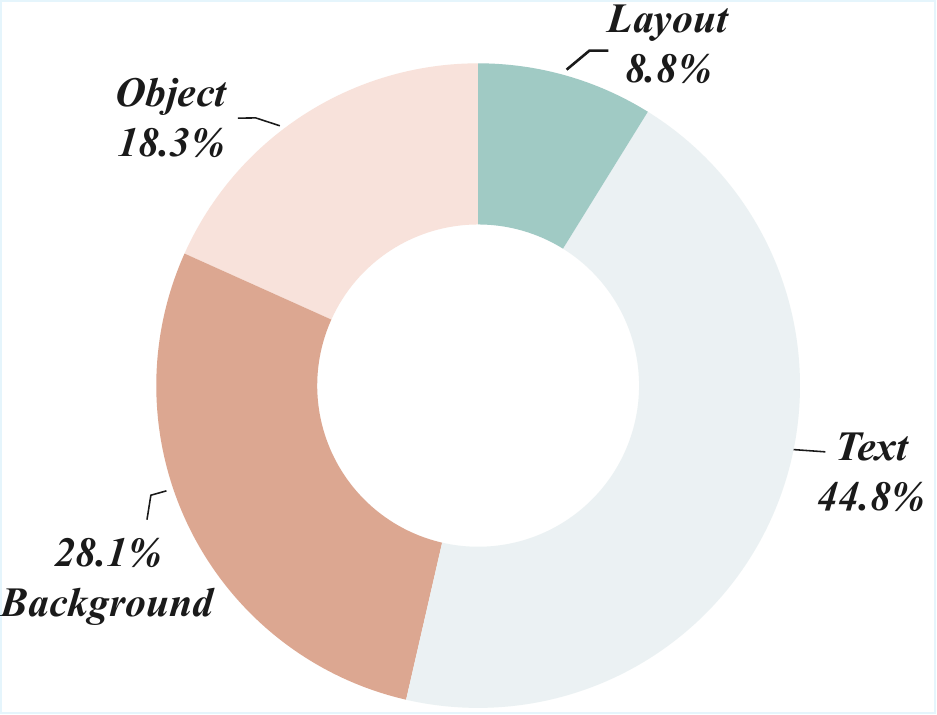} 
    \subcaption{`Weakest Link' Analysis}
    \label{fig:stats_weakest}
\end{subfigure}

\caption{\textbf{Statistical Profile of E-comIQ-18k.} 
    (a) The multi-modal distribution of overall scores highlights sample diversity. 
    (b) The distribution of CoT rationale lengths .
    (c) The correlation matrix reveals a semi-orthogonal dimensional structure.
    (d) A `weakest link' analysis pinpoints common diagnostic challenges.
}
\label{fig:statistics_panel} 

\end{figure}

The dataset’s statistical properties, summarized in Fig.~\ref{fig:statistics_panel}, support its suitability for fine grained diagnostic assessment. We observe a broad, multimodal score distribution across the four dimensions (Fig.~\ref{fig:statistics_panel}a) and long CoT rationales with an average length above 800 Chinese characters (Fig.~\ref{fig:statistics_panel}b). The four dimensions are only weakly correlated, with a mean interdimensional Pearson correlation of $\rho \approx 0.24$ (Fig.~\ref{fig:statistics_panel}c), indicating that a single holistic score is insufficient to capture e-commerce poster quality. A weakest link analysis over images with any dimension below 3.0 shows that Text is the bottleneck in 44.8\% of cases (Fig.~\ref{fig:statistics_panel}d) and also has the strongest correlation with overall quality ($\rho = 0.67$), highlighting the central role of text quality in Chinese e-commerce posters. Additional statistics on checklist tags, source distributions, and annotator variance are provided in the Appendix.
%

\begin{table*}[t!]
\centering
\caption{
    \textbf{Correlation performance against state-of-the-art models on the E-comIQ-18k test set.} 
    Each cell reports \textbf{PLCC / SRCC}. The \textbf{best} result is in \textbf{bold}, and the \textit{second-best} is \underline{underlined}.
}
\label{tab:corr_results}
\setlength{\tabcolsep}{9pt} 
\footnotesize
\begin{tabular}{l|cc|cc|cc|cc|cc}
\toprule
\textbf{Model} &
\multicolumn{2}{c|}{\textbf{Overall}} &
\multicolumn{2}{c|}{\textbf{Background}} &
\multicolumn{2}{c|}{\textbf{Object}} &
\multicolumn{2}{c|}{\textbf{Text}} &
\multicolumn{2}{c}{\textbf{Layout}} \\
& \textit{PLCC} & \textit{SRCC} &
\textit{PLCC} & \textit{SRCC} &
\textit{PLCC} & \textit{SRCC} &
\textit{PLCC} & \textit{SRCC} &
\textit{PLCC} & \textit{SRCC} \\
\midrule
\rowcolor{gray!10}
\multicolumn{11}{l}{\textit{Traditional NR-IQA Models}} \\
MUSIQ~\cite{ke2021musiq} 
& 0.074 & 0.081 & 0.143 & 0.158 & 0.064 & 0.066 & 0.026 & 0.066 & 0.074 & 0.081 \\
SPAQ~\cite{fang2020perceptual}  
& -0.174 & -0.172 & -0.271 & -0.279 & -0.069 & -0.119 & -0.047 & -0.046 & -0.161 & -0.167 \\

\midrule
\rowcolor{gray!10}
\multicolumn{11}{l}{\textit{General-Purpose MLLMs}} \\
GPT-4o~\cite{openai2023gpt}  
& 0.242 & 0.219 & 0.368 & 0.413 & 0.105 & 0.122 & 0.126 & 0.148 & 0.297 & 0.282 \\
Gemini 2.5 Pro~\cite{comanici2025gemini} 
& 0.213 & 0.228 & 0.334 & 0.419 & 0.101 & 0.098 & 0.146 & 0.122 & 0.350 & 0.320 \\
Claude-Sonnet-4.5~\cite{claude45} 
& 0.213 & 0.228 & 0.334 & 0.419 & 0.101 & 0.098 & 0.146 & 0.122 & 0.350 & 0.320 \\
Grok-4~\cite{xai2024grok} 
& 0.178 & 0.150 & 0.299 & 0.356 & 0.138 & 0.105 & 0.113 & 0.118 & 0.267 & 0.250 \\
Qwen2.5-VL-72B~\cite{bai2025qwen2} 
& 0.127 & 0.144 & 0.281 & 0.308 & -0.028 & -0.057 & 0.100 & 0.070 & 0.142 & 0.153 \\
Qwen2.5-VL-7B~\cite{bai2025qwen2} 
& 0.035 & 0.119 & 0.206 & 0.222 & 0.093 & 0.075 & 0.040 & 0.042 & 0.167 & 0.134 \\
Qwen3-VL-8B~\cite{bai2025qwen2} 
& 0.212 & 0.195 & 0.328 & 0.368 & 0.098 & 0.110 & 0.137 & 0.148 & 0.231 & 0.204 \\

\midrule
\rowcolor{gray!10}
\multicolumn{11}{l}{\textit{Specialized Evaluators}} \\
C2Score~\cite{zhu2024adaptive} 
& 0.158 & 0.171 & 0.149 & 0.174 & 0.147 & 0.152 & 0.084 & 0.149 & 0.127 & 0.142 \\
Q-Align~\cite{wu2023q} 
& 0.188 & 0.182 & 0.327 & 0.355 & 0.135 & 0.153 & -0.001 & 0.001 & 0.112 & 0.109 \\
VQ-R1~\cite{wu2025visualquality} 
& 0.227 & 0.257 & 0.152 & 0.233 & 0.213 & 0.222 & 0.112 & 0.134 & 0.251 & 0.265 \\
DeQA~\cite{han2025deqa} 
& 0.193 & 0.189 & 0.214 & 0.236 & 0.191 & 0.209 & 0.048 & 0.054 & 0.103 & 0.107 \\
Q-Insight~\cite{li2025q} 
& 0.183 & 0.152 & 0.231 & 0.251 & -0.032 & -0.071 & -0.024 & -0.027 & 0.134 & 0.149 \\

\midrule
\rowcolor{gray!10}
\multicolumn{11}{l}{\textit{Fine-tuned Models}} \\
Q-Insight+GRPO~\cite{li2025q} 
& 0.265 & 0.235 & 0.312 & 0.312 & 0.123 & 0.070 & 0.096 & 0.132 & 0.221 & 0.218 \\
Q-Insight+STF & 0.297 & 0.319 & 0.442 & 0.478 & 0.242 & 0.244 & 0.291 & 0.304 & 0.379 & 0.391 \\
Q-Insight+STF+GRPO    & 0.338 & 0.348 & 0.459 & 0.496 & 0.386 & 0.304 & 0.320 & 0.342 & 0.375 & 0.403 \\
Qwen2.5-VL-7B+SFT~\cite{bai2025qwen2} 
& \underline{0.346} & \underline{0.346} & \underline{0.458} & \textbf{0.530} & \underline{0.272} & \underline{0.238} & \underline{0.272} & \underline{0.283} & \underline{0.390} & \underline{0.418}\\
\textbf{E-comIQ-M (Ours)} 
& \textbf{0.425} & \textbf{0.433} & \textbf{0.496} & \underline{0.520} & \textbf{0.391} & \textbf{0.361} & \textbf{0.364} & \textbf{0.392} & \textbf{0.483} & \textbf{0.506} \\

\bottomrule
\end{tabular}
\end{table*}

\section{E-comIQ-M}
\label{sec:method}

\subsection{Training Strategy}
\label{subsec:methodology}

E-comIQ-M is implemented by fine-tuning a multimodal language model to act as an e-commerce poster evaluator. Given an input image and an evaluation instruction, the model outputs a structured JSON object containing four dimension scores (Object, Background, Text, Layout) and an overall score, together with an optional natural language rationale. We adopt \textbf{Qwen-2.5-VL-7B}~\cite{bai2025qwen2} as the backbone due to its strong vision and language capabilities and native support for Chinese. The training procedure consists of two stages: SFT on the full 15k training set to learn domain knowledge and output format, followed by GRPO on a curated hard subset to refine score calibration.

\vspace{-10pt}
\paragraph{Stage 1: SFT.}
We conduct SFT on the entire 15k training set, using the expert scores and CoT rationales as targets. This stage teaches the model the task format, domain specific concepts, and a reasonable initial scoring behavior.
\vspace{-10pt}
\paragraph{Stage 2: GRPO.}
In this stage, we optimize the policy $\pi_{\theta}$ with GRPO~\cite{guo2025deepseek} on a curated hard subset $\mathcal{D}_{\text{hard}}$ of 3k training samples, obtained by ranking all 15k examples by the SFT model's mean squared error (MSE) and retaining the worst 3k. The objective is to maximize the expectation of a final reward $R(x,y)$, defined as

\begin{equation}
    R(x, y) = R_{\text{score}}(x, y) + \lambda_{\text{fmt}} R_{\text{fmt}}(y),
\label{eq:final_reward}
\end{equation}
where $R_{\text{fmt}}(y)$ is a binary reward that equals 1 if the output can be parsed as a valid JSON object and 0 otherwise, and $\lambda_{\text{fmt}}$ balances the score and format rewards.

The score reward $R_{\text{score}}$ is a convex combination of an accuracy component and a distribution component:
\begin{equation}
    R_{\text{score}}(x, y) = \lambda_{\text{score}} R_{\text{acc}}(x, y) + (1 - \lambda_{\text{score}}) R_{\text{dist}}(x, y),
\end{equation}
where the trade-off hyperparameter $\lambda_{\text{score}}$ is empirically set to 0.65. The two components are defined as follows.

\begin{itemize}
    \item \textit{Accuracy reward $R_{\text{acc}}$.}
    Let $S_{\text{pred}}^i(y)$ and $S_{\text{gt}}^i$ denote the predicted and ground-truth scores for the $i$-th dimension, where $i\in\{1,\dots,5\}$ indexes the four functional dimensions and the overall score. We define
    \begin{equation}
        R_{\text{acc}}(x, y) = \frac{1}{5} \sum_{i=1}^{5} p_i \cdot \mathbb{1}\big(|S_{\text{pred}}^i(y) - S_{\text{gt}}^i| \le \tau \big),
    \end{equation}
    where $\mathbb{1}(\cdot)$ is the indicator function, $\tau = 0.2$, and the penalty factor $p_i$ down-weights predictions that cross expert-defined quality tiers:
    \[
        p_i =
        \begin{cases}
            0.7, & \text{if } \mathrm{tier}\!\left(S_{\text{pred}}^i(y)\right) \neq \mathrm{tier}\!\left(S_{\text{gt}}^i\right), \\
            1.0, & \text{otherwise}.
        \end{cases}
    \]
    Here $\mathrm{tier}(\cdot)$ maps a score to one of the three quality levels (poor, good, excellent) defined in Sec.~\ref{subsec:annotation}.

    \item \textit{Distribution reward $R_{\text{dist}}$.}
    Let $\vec{v}_{\text{pred}}(y), \vec{v}_{\text{gt}} \in \mathbb{R}^4$ denote the 4D sub-score vectors over Object, Background, Text, and Layout. We measure geometric consistency via an exponential penalty on their Euclidean distance:
    \begin{equation}
        R_{\text{dist}}(x, y) = \exp\left(-\alpha \cdot \big\|\vec{v}_{\text{pred}}(y) - \vec{v}_{\text{gt}}\big\|_2\right),
    \end{equation}
    where the scaling hyperparameter $\alpha$ is set to 0.5.
\end{itemize}

\begin{table*}[t!]
\centering
\caption{
    \textbf{Accuracy performance against state-of-the-art models on the E-comIQ-18k test set.} 
    Each cell reports \textbf{Acc@0.5 / Acc@1.0 (in \%)}. The \textbf{best} result is in \textbf{bold}, and the \textit{second-best} is \underline{underlined}.
}
\label{tab:acc_results}
\setlength{\tabcolsep}{5pt} 
\footnotesize

\begin{tabular}{l|cc|cc|cc|cc|cc}
\toprule
\textbf{Model} &
\multicolumn{2}{c|}{\textbf{Overall}} &
\multicolumn{2}{c|}{\textbf{Background}} &
\multicolumn{2}{c|}{\textbf{Object}} &
\multicolumn{2}{c|}{\textbf{Text}} &
\multicolumn{2}{c}{\textbf{Layout}} \\
& \textit{Acc@0.5} & \textit{Acc@1.0} &
\textit{Acc@0.5} & \textit{Acc@1.0} &
\textit{Acc@0.5} & \textit{Acc@1.0} &
\textit{Acc@0.5} & \textit{Acc@1.0} &
\textit{Acc@0.5} & \textit{Acc@1.0} \\
\midrule
\rowcolor{gray!10}
\multicolumn{11}{l}{\textit{General-Purpose MLLMs}} \\
GPT-4o~\cite{openai2023gpt}  
& 32.4 & 59.0 & 45.8 & 70.4 & 47.8 & 70.8 & 34.3 & 55.0 & 48.8 & 74.8 \\
Gemini 2.5 Pro~\cite{comanici2025gemini} 
& 26.6 & 51.0 & 50.4 & 73.8 & 42.2 & 63.4 & 29.2 & 45.0 & 46.0 & 68.0 \\
Claude-Sonnet-4.5~\cite{claude45} 
& 28.4 & 57.2 & 53.4 & 77.8 & 49.4 & 71.4 & 27.8 & 45.2 & 50.1 & 74.8 \\
Grok-4~\cite{xai2024grok} 
& 33.3 & 57.3 & 46.3 & 69.5 & 42.9 & 65.7 & 26.9 & 41.1 & 43.9 & 73.0 \\
Qwen2.5-VL-72B~\cite{bai2025qwen2} 
& 18.6 & 40.0 & 44.2 & 68.2 & 42.6 & 65.2 & 25.2 & 42.0 & 39.2 & 58.2 \\
Qwen2.5-VL-7B~\cite{bai2025qwen2} 
& 29.3 & 53.7 & 40.0 & 68.0 & 40.2 & 67.6 & 30.1 & 48.8 & 35.9 & 67.8 \\
Qwen3-VL-8B~\cite{bai2025qwen2} 
& 26.0 & 48.0 & 37.8 & 63.0 & 44.4 & 65.6 & 26.8 & 39.2 & 39.8 & 55.8 \\

\midrule
\rowcolor{gray!10}

\multicolumn{11}{l}{\textit{Specialized Evaluators}} \\
Q-Insight~\cite{li2025q} 
& 43.8 & 72.6 & 51.0 & \textbf{85.4} & 45.8 & \underline{81.0} & 34.0 & 69.4 & 40.0 & \textbf{86.6} \\
VQ-R1~\cite{wu2025visualquality} 
& 13.6 & 35.0  & 30.0 & 53.8 & 48.8 & 71.4 & 23.4 & 39.8  & 38.6 & 58.6  \\
Q-Align~\cite{wu2023q} 
& 30.4 & 57.2  
& 38.2 & 72.6  
& 49.6 & 76.6  
& 19.6 & 57.2  
& 30.0 & 74.4  \\

DeQA~\cite{han2025deqa} 
& 39.4 & 64.6  
& 39.2 & 77.6  
& 40.8 & 75.8  
& 28.4 & 62.6  
& 33.4 & 76.4  \\
\midrule
\rowcolor{gray!10}

\multicolumn{11}{l}{\textit{Fine-tuned Models}} \\
Q-Insight+GRPO~\cite{li2025q} 
& 47.0 & 76.8 & 46.8 & 79.8 & 47.6 & \textbf{81.2} & 34.2 & \underline{70.8} & 39.6 & 81.4 \\

Q-Insight+STF & 50.8 & 75.2 & 63.0 & 81.6 & 49.4 & 74.4 & 43.8 & 67.4 & 55.2 & 77.8 \\
Q-Insight+STF+GRPO  & 53.6 & 79.6 & 60.2 & 78.4 & 51.4 & 74.0 & 47.2 & 68.4 & 57.8 & 77.2 \\

Qwen2.5-VL-7B+SFT~\cite{bai2025qwen2} 
& \underline{51.0} & \underline{78.8} & \underline{63.2} & 79.2 & \underline{51.2} & 75.4 & \underline{43.8} & 69.6 & \underline{57.8} & 81.0 \\
\textbf{E-comIQ-M (Ours)} 
& \textbf{55.6} & \textbf{81.8} & \textbf{65.0} & \underline{81.6} & \textbf{51.4} & 78.2 & \textbf{49.6} & \textbf{75.0} & \textbf{63.2} & \underline{83.0} \\
\bottomrule
\end{tabular}
\end{table*}

\begin{table*}[t!]
\centering
\caption{
    \textbf{Ablation studies on the E-comIQ-18k test set.} 
    We analyze the impact of different training stages and reward function designs. 
    Each cell reports \textbf{PLCC / SRCC}.
}
\label{tab:ablation_study}
\setlength{\tabcolsep}{9.5pt} 
\footnotesize

\begin{tabular}{l|cc|cc|cc|cc|cc}
\toprule
\textbf{Config.} &
\multicolumn{2}{c|}{\textbf{Overall}} &
\multicolumn{2}{c|}{\textbf{Background}} &
\multicolumn{2}{c|}{\textbf{Object}} &
\multicolumn{2}{c|}{\textbf{Text}} &
\multicolumn{2}{c}{\textbf{Layout}} \\
& \textit{PLCC} & \textit{SRCC} &
\textit{PLCC} & \textit{SRCC} &
\textit{PLCC} & \textit{SRCC} &
\textit{PLCC} & \textit{SRCC} &
\textit{PLCC} & \textit{SRCC} \\
\midrule
GRPO only 
& 0.158 & 0.154 & 0.174 & 0.218 & 0.013 & 0.022 & 0.059 & 0.077 & 0.228 & 0.192 \\
SFT only 
& 0.346 & 0.346 & 0.458 & \underline{0.530} & 0.272 & 0.238 & 0.272 & 0.283 & 0.390 & 0.418 \\
SFT+GRPO (Simple) 
& \underline{0.352} & \underline{0.360} & \underline{0.466 }& 0.509 & \underline{0.287} & \underline{0.261} & \underline{0.368} & \underline{0.385} & \underline{0.413} & \underline{0.445} \\
\textbf{SFT+GRPO (Complex)} 
& \textbf{0.413} & \textbf{0.410} & \textbf{0.509} & \textbf{0.540} & \textbf{0.320} & \textbf{0.274} & \textbf{0.391} & \textbf{0.405} & \textbf{0.457} & \textbf{0.466} \\
\bottomrule
\end{tabular}
\end{table*}



\subsection{Experimental Evaluation}
\label{subsec:evaluation}

\paragraph{Setup.}

We evaluate all models on the 1k test set using Pearson (PLCC) and Spearman (SRCC) correlations, together with absolute accuracy Acc@$k$. Here Acc@$k$ denotes the percentage of predictions whose absolute error against the ground truth is at most $k$, and we report $k=0.5$ and $k=1.0$. We compare our method against four groups of baselines: (1) traditional NR-IQA models, including MUSIQ~\cite{ke2021musiq} and SPAQ~\cite{fang2020perceptual}; (2) general-purpose MLLMs, such as GPT-4o~\cite{openai2023gpt}, Gemini 2.5 Pro~\cite{comanici2025gemini}, and the Qwen family~\cite{bai2025qwen2}; (3) specialized evaluators, including C2Score~\cite{zhu2024adaptive}, Q-Align~\cite{wu2023q}, VQ-R1~\cite{wu2025visualquality}, DeQA~\cite{han2025deqa}, and Q-Insight~\cite{li2025q}; and (4) fine-tuned models, namely Q-Insight+GRPO~\cite{li2025q}, Qwen2.5-VL-7B+SFT, and our E-comIQ-M. For models whose output scores are not normalized to our $[1,5]$ scale (all traditional NR-IQA models and some specialized evaluators such as C2Score), we only report correlation metrics and omit Acc@$k$.

\begin{figure*}[t]
    \centering

    \includegraphics[width=\linewidth]{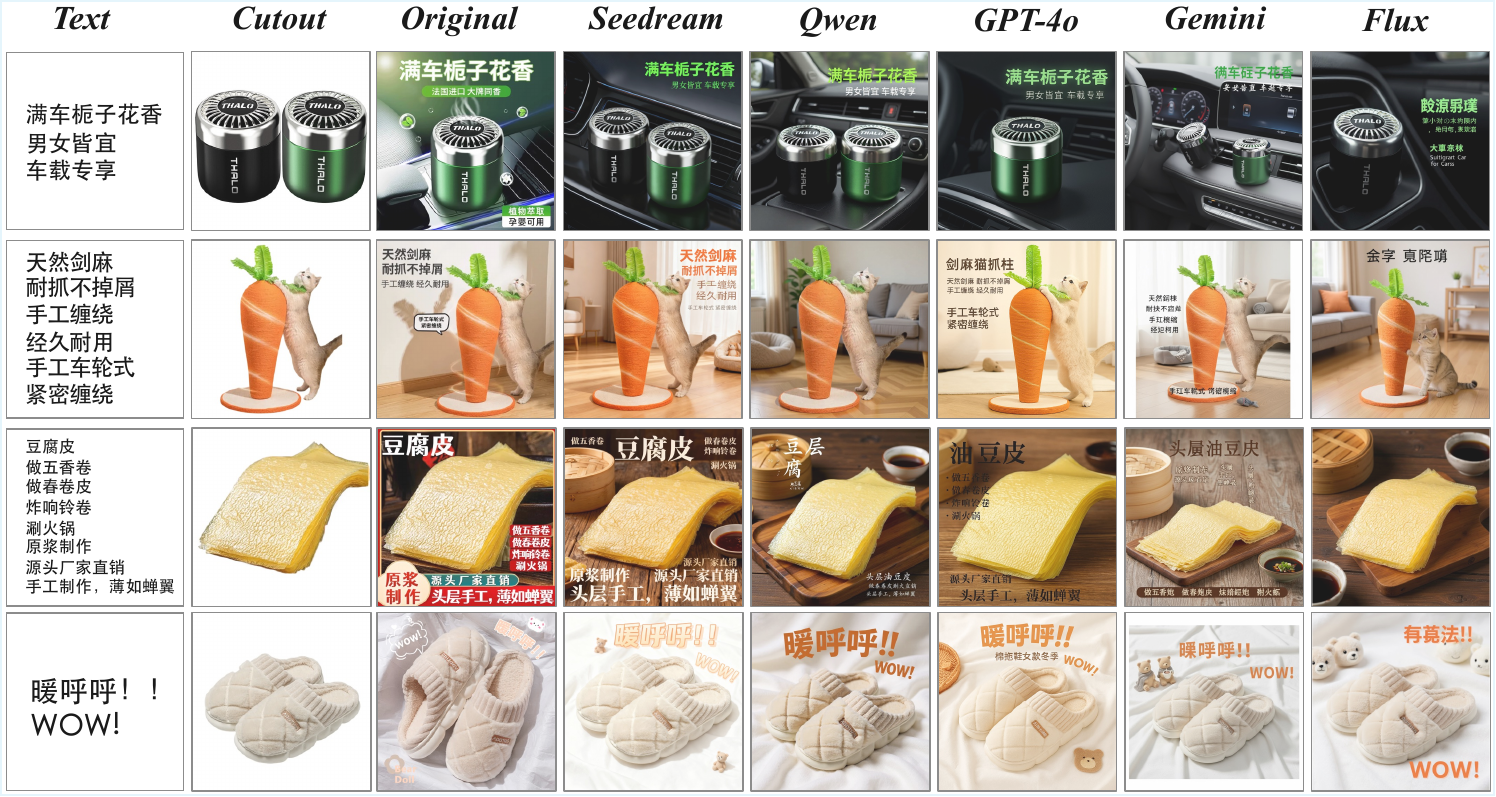}
    \caption{ Qualitative comparison of images generated by leading models on E-comIQ-Bench. 
    }
    \label{fig:example}
\end{figure*}

\vspace{-10pt}
\paragraph{Main results.}
\begin{itemize}\itemsep0pt
    \item \textbf{Inadequacy of existing models.}
 In E-comIQ-18k, most baseline models obtain overall SRCC values below 0.3 (Table~\ref{tab:corr_results}) and relatively low Acc@0.5 (Table~\ref{tab:acc_results}). They are consistently stronger on background dimension than text, which suggests that these models mainly rely on simple global cues and generic aesthetic knowledge. In contrast, Chinese e-commerce posters require fine grained domain specific reasoning about dense characters, copy correctness, readability, and how text and products jointly convey core selling points, and these aspects are largely absent from the objectives and training data of existing models. This gap indicates that simply applying existing IQA or AIGC models is not sufficient and motivates both a dedicated evaluator and a curated in domain dataset that explicitly encodes these e-commerce specific criteria.

    \item \textbf{Effect of domain-specific SFT.}
    For Qwen2.5-VL-7B~\cite{bai2025qwen2}, overall SRCC increases from 0.119 to 0.346 (Table~\ref{tab:corr_results}), and overall Acc@0.5 rises from 29.3\% to 51.0\% (Table~\ref{tab:acc_results}), with especially strong improvements on the text and layout dimensions. After SFT, the model already surpasses existing baselines on most metrics, which shows that our dataset provides effective supervision for the domain specific criteria that other models do not learn well, rather than only reinforcing generic aesthetic preferences.

    \item \textbf{Benefit of the two stage strategy.}
    
    Building on the SFT model, our full evaluator E-comIQ-M further improves both correlation and accuracy. Overall SRCC increases from 0.346 to 0.433 and overall Acc@0.5 from 51.0\% to 55.6\%, with the largest gains again on text and layout. By contrast, extending Q-Insight with GRPO under its original reward design, which optimizes scores without using CoT information, brings only limited gains on our benchmark, reflecting the difficulty of learning stable score distributions in this task from weak signals alone. Taken together, these results show that the two stage SFT plus GRPO strategy can refine score calibration beyond supervised learning alone and makes E-comIQ-M a reliable automated evaluator for e-commerce poster quality.
\end{itemize}

\vspace{-10pt}
\paragraph{Ablation Studies.}

As shown in Table~\ref{tab:ablation_study}, we compare four configurations: (a) \textit{GRPO only}, which trains Qwen2.5-VL-7B without SFT; (b) \textit{SFT only}, the baseline model after supervised fine tuning; (c) \textit{SFT+GRPO (Simple)}, which uses a reward composed only of the accuracy term $R_{\text{acc}}$; and (d) \textit{SFT+GRPO (Complex)}, our full model with both accuracy and distribution terms $R_{\text{acc}} + R_{\text{dist}}$. GRPO only is clearly worse than SFT only, indicating that reinforcement learning alone is not sufficient for this multi dimensional continuous scoring task. Starting from the SFT checkpoint, adding GRPO with the simple accuracy reward improves performance, especially on text, which shows that preference optimisation on hard samples helps correct systematic biases. The full SFT+GRPO (Complex) configuration achieves the best overall correlations and accuracies, suggesting that the distribution term further aligns the geometry of sub scores with expert judgement. Sensitivity experiments on reward weights and hard subset size in the Appendix confirm that these trends are robust.

\section{E-comIQ-Bench}
\label{sec:benchmark}


\begin{table*}[t!]
\centering
\caption{
    \textbf{Benchmark results for leading generative E-comIQ-Ms on E-comIQ-Bench.} 
}
\label{tab:benchmark_results}
\setlength{\tabcolsep}{6pt}
\footnotesize
\begin{tabular}{l|cc|cc|cc|cc|cc}
\toprule
\textbf{E-comIQ-M} &
\multicolumn{2}{c|}{\textbf{Overall}} &
\multicolumn{2}{c|}{\textbf{Background}} &
\multicolumn{2}{c|}{\textbf{Object}} &
\multicolumn{2}{c|}{\textbf{Text}} &
\multicolumn{2}{c}{\textbf{Layout}} 
 \\
& \textit{Human} & \textit{E-comIQ-M} &
\textit{Human} & \textit{E-comIQ-M} &
\textit{Human} & \textit{E-comIQ-M} &
\textit{Human} & \textit{E-comIQ-M} &
\textit{Human} & \textit{E-comIQ-M} \\
\midrule
SeeDream~\cite{seedream2025seedream} & \textbf{3.65} & \textbf{3.53} & 4.70 & 4.40 & 3.78 & 4.08 & \textbf{4.05} & \textbf{3.98} & 4.51 & 4.37 \\
Qwen~\cite{wu2025qwen}  & 3.26 & 3.36 & \textbf{4.71} & \textbf{4.51} & \textbf{3.84} & 4.11 & 3.17 & 2.87 & \textbf{4.55} & 4.27 \\
GPT-4o~\cite{openai2024sora}     & 2.76 & 2.76 & 4.67 & 4.19 & 3.49 & \textbf{4.12} & 2.83 & 3.49 & 4.36 & 4.36 \\
Gemini~\cite{comanici2025gemini}   & 1.92 & 2.91 & 4.65 & 4.26 & 3.72 & 3.73 & 1.45 & 2.41 & 4.52 & 4.30 \\
Flux ~\cite{batifol2025flux}    & 1.89 & 2.72 & 4.60 & 4.19 & 3.66 & 3.76 & 1.48 & 1.98 & 4.56 & \textbf{4.40}  \\
\midrule
Original & 3.78 & 3.78 & 3.84 & 3.84 & 4.40 & 4.01 & 4.01 & 3.31 & 4.02 & 3.82  \\
\bottomrule
\end{tabular}
\end{table*}
 
\subsection{Design and Protocol}
\label{subsec:benchmark_design}

E-comIQ-Bench contains 500 test cases, each with a product foreground cutout, its original merchant poster, and a Chinese prompt. The products cover seven major e-commerce categories (Fig.~\ref{fig:donut}). Prompts are constructed from the product’s key selling points extracted from the original listing, then rewritten with MLLM assistance and lightly edited by the authors to ensure suitability for poster generation. For each cutout–prompt pair, we query several leading text-to-image systems and obtain one generated poster per model; the original merchant poster serves as a human-designed reference.

\textbf{Core quality assessment.}
A panel of professional e-commerce designers rates every poster along the five dimensions used in E-comIQ-18k (\textit{Overall, Object, Background, Text, Layout}). In parallel, our evaluator E-comIQ-M predicts the same five scores from the image, providing a scalable automatic benchmark that can be compared against human judgement.

\textbf{Auxiliary diagnostic metrics.}
E-comIQ-M is a non-reference quality model, we complement it with reference-based diagnostics computed against the original poster. We measure \textit{subject fidelity} using DINO similarity, LPIPS distance and CLIP score between the product region in the original and generated images, and \textit{text content accuracy} using phrase-level F1 and character-level normalised Levenshtein similarity between the prompt text and OCR-extracted text. Implementation details and the full metric configuration are provided in the Appendix, together with an open-source evaluation toolbox.




\subsection{Results and Diagnostic Analysis}
\label{subsec:benchmark_results}
\begin{figure}[h]
\centering
\begin{subfigure}[b]{0.45\linewidth}
    \centering
    \includegraphics[width=\linewidth]{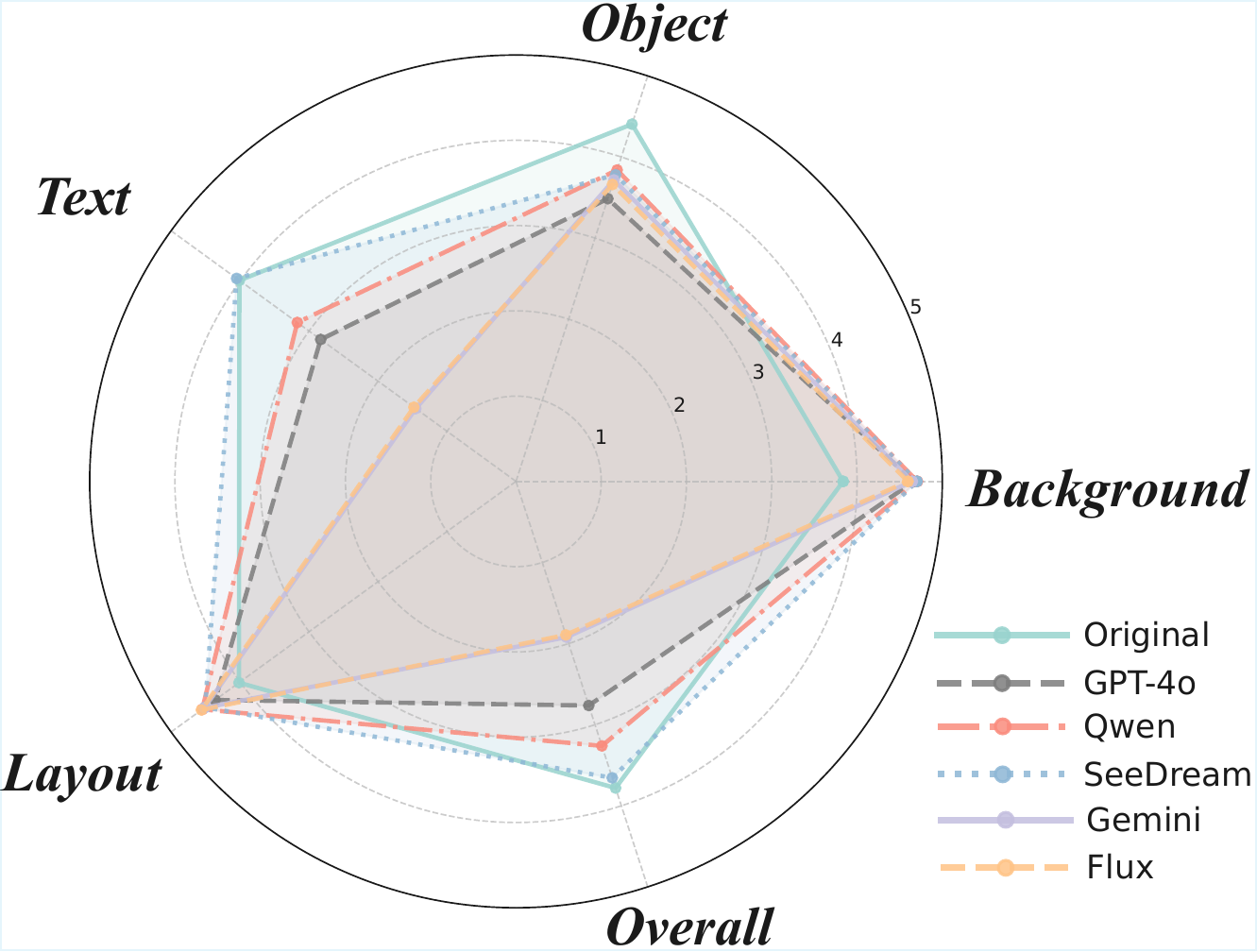}
    \caption{}
    \label{fig:radar}
\end{subfigure}%
\hfill
\begin{subfigure}[b]{0.51\linewidth}
    \centering
    \includegraphics[width=\linewidth]{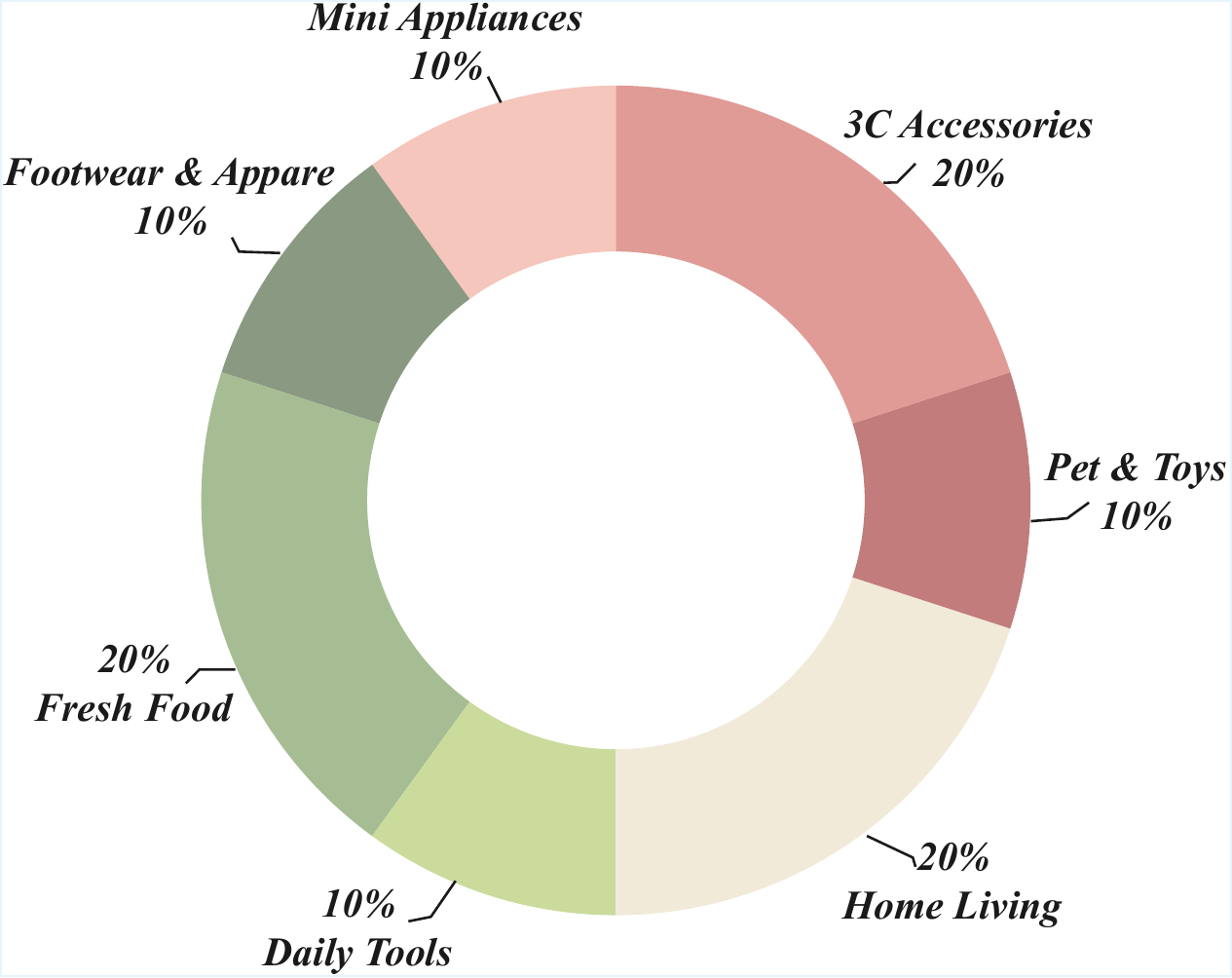}
    \caption{}
    \label{fig:donut}
\end{subfigure}
\caption{
    (a) Multi-model performance comparison across five dimensions via radar chart. 
    (b) Distribution of image categories.
}
\label{fig:benchmark_analysis}
\end{figure}

\paragraph{Overall performance.}
Table~\ref{tab:benchmark_results} and Fig.~\ref{fig:radar} report human scores for all models on E-comIQ-Bench. The strongest generative model slightly surpasses the average quality of original merchant posters on the overall dimension, while other systems still lag behind, which suggests that current text-to-image models are close to but do not clearly exceed typical human designs. Across dimensions, backgrounds and layouts are often rated higher than those of original posters, whereas text and, to a lesser extent, object quality remain the main bottlenecks.
\vspace{-10pt}
\paragraph{Consistency between human and E-comIQ-M.}
We next examine how well E-comIQ-M tracks human judgements on this benchmark. Although the overall PLCC and SRCC between E-comIQ-M and human scores are only around 0.34 in this challenging out-of-domain setting, the model reproduces the relative ranking of systems and the dimension-wise strength profiles in Table~\ref{tab:benchmark_results} and Fig.~\ref{fig:radar} reasonably well. In particular, models that human judge as strong or weak on the text dimension receive systematically higher or lower text scores from E-comIQ-M, which supports using it as a scalable automatic evaluator with human scores as the primary reference.
\vspace{-10pt}
\paragraph{Insights from auxiliary metrics.}
Table~\ref{tab:objective_metrics} analyses the auxiliary reference-based metrics. Subject fidelity indicators between the product regions of the original and generated posters broadly follow the human and E-comIQ-M object scores, showing that they are useful for detecting mismatched subjects. In contrast, text content accuracy often disagrees with human and E-comIQ-M text scores: some models with low text ratings, such as GPT-4o and Gemini, still achieve high OCR-based phrase and character similarity. Manual inspection reveals many subtle stroke errors and visually incorrect characters that OCR systems nonetheless map to the intended phrase, making these posters unacceptable in a commercial setting. This mismatch shows that OCR-style metrics are unreliable for evaluating Chinese text rendering and highlights the value of E-comIQ-M, which is explicitly aligned with human text judgements.

\begin{table}[h!]
\centering
\caption{
    \textbf{Objective metrics for subject fidelity and text content accuracy on E-comIQ-Bench.}
}
\label{tab:objective_metrics}
\footnotesize
\setlength{\tabcolsep}{1pt}
\begin{tabular}{l|ccc|cc}
\toprule
\textbf{Model} &
\multicolumn{3}{c|}{\textbf{Subject Fidelity}} &
\multicolumn{2}{c}{\textbf{Text Content Accuracy}} \\
& \textit{DINO Sim $\uparrow$} & \textit{LPIPS $\downarrow$} & \textit{CLIP Score $\uparrow$} &
\textit{Phrase F1 $\uparrow$} & \textit{Char Sim $\uparrow$} \\
\midrule
SeeDream~\cite{seedream2025seedream} & 0.74 & \underline{0.62} & 0.81 & \textbf{0.88} & \textbf{0.92}  \\
Qwen~\cite{wu2025qwen}  & \textbf{0.81} & \textbf{0.57} & \textbf{0.84} & 0.49 & 0.54 \\
GPT-4o~\cite{openai2024sora}     & 0.73 & 0.67 & 0.81 & \underline{0.86} & \underline{0.91} \\
Gemini~\cite{comanici2025gemini}   & 0.67 & 0.69 & 0.78 & 0.56 & 0.62 \\
Flux~\cite{batifol2025flux}     & \underline{0.76} & 0.64 & \underline{0.82} & 0.10 & 0.10 \\
\bottomrule
\end{tabular}
\end{table}

\section{Conclusion}

We introduced a domain-specific framework for Chinese e-commerce IQA, including a multi-dimensional quality standard, the E-comIQ-18k dataset with expert scores and CoT rationales, and the E-comIQ-Bench benchmark with an automatic evaluation toolbox. Based on these resources, our evaluator E-comIQ-M aligns better with expert judgements than general-purpose models and supports large-scale, fine-grained analysis of text-to-image systems in realistic commercial settings. At the same time, important limitations remain: as a non-reference model, E-comIQ-M cannot directly measure subject identity fidelity, and its PLCC/SRCC with human scores, while improved, are still moderate on challenging out-of-domain data. These gaps show that robust, human-aligned evaluation for commercial AIGC is still an open and promising direction. We plan to explore stronger calibration strategies, and hope that our works will serve as useful building blocks for future research.


{
    \small
    \bibliographystyle{ieeenat_fullname}
    \bibliography{main}
}
\clearpage
\maketitlesupplementary

\renewcommand{\thesection}{A}
\section{Dataset: E-comIQ-18k Details}
\label{sec:rationale}

\subsection{Source Composition and Splits}       
\label{app:source_composition}
E-comIQ-18k contains 18k images drawn from six sources  (See Figure 4 in Sec. 3.1). The proportions are
27.8\% merchant HQ, 27.8\% merchant LQ, 16.7\% open-source posters,
11.1\% AI-generated posters, 11.1\% AI-edited posters, and 5.6\% professional designs.
\vspace{-10pt}
\paragraph{Merchant originals (HQ / LQ).}
We start from a large pool of merchant-provided product photos collected from
real online listings. Each image is labelled by experts with a binary
High Quality(HQ) / Low Quality(LQ) according to overall commercial usability, including product
visibility, background cleanliness, text legibility, and layout. Then we randomly sample 5k HQ and 5k
LQ images, removing obvious near-duplicates. This procedure gives a broad and
realistic quality spectrum for in-the-wild merchant content.

\vspace{-10pt}
\paragraph{Open-source posters.}
To increase diversity in style and category coverage, we further sample
3k posters from a public e-commerce poster dataset released in Autoposter~\cite{lin2023autoposter}. These images are usually complete posters with designed

\vspace{-10pt}
\paragraph{AI-generated posters.}
The AI-generated subset is created from product cutouts on a white
background. For each product we construct a text prompt that specifies the
scene, style, and key selling points, then use GPT-4o as a text-to-image
generator conditioned on the cutout as visual reference. Generation prompts
and examples are provided in Fig ~\ref{fig:soracase}, and we discard obvious failures
such as missing products or unreadable text.

\begin{figure}[htbp]
    \centering
    \includegraphics[width=\linewidth]{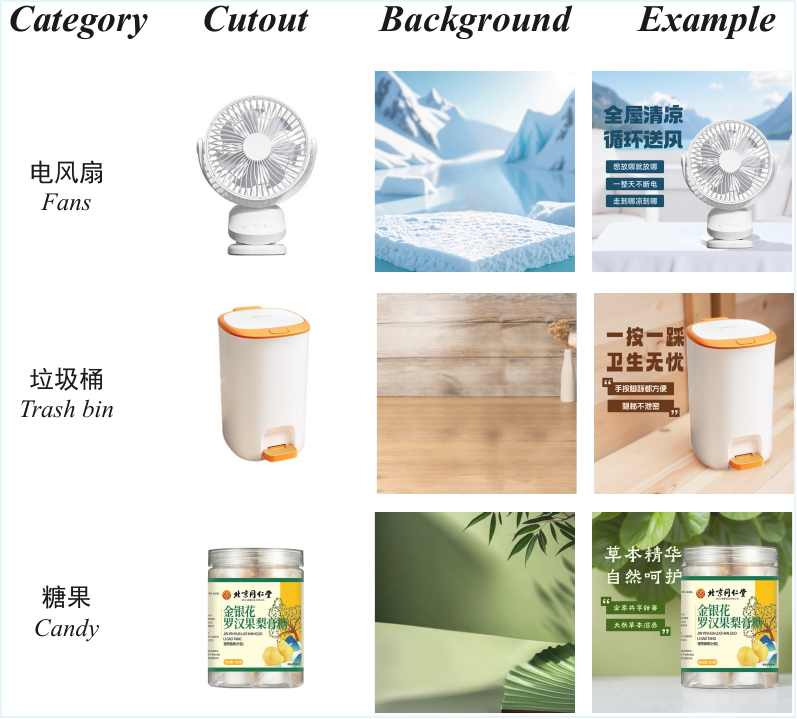}
    \caption{\textbf{AI-edited posters via Flux.}
Given a product cutout and its category (left), we retrieve a matching scene
from a predefined background library (middle) and feed both into Flux to
compose a subject–background image. The final poster (right) is obtained by
adding Chinese marketing copy using predefined text templates.
    }
    \label{fig:fluxcase}
\end{figure}

\vspace{-10pt}
\paragraph{AI-edited posters.}
The AI-edit subset is created by a multi-stage automatic pipeline shown in Fig~\ref{fig:fluxcase} that mimics
template-based design. Given a product cutout on a white background and its
category, we first retrieve a compatible scene from a predefined background
library. The cutout and selected background are then jointly fed into Flux to
generate a composed image with the subject placed in context. Finally, we
render Chinese marketing copy into predefined text templates according to
handcrafted layout rules.

\vspace{-10pt}
\paragraph{Professional designs.}
The professional subset contains posters manually crafted by experienced
e-commerce designers using standard design software.

\begin{figure}[htbp]
    \centering
    \includegraphics[width=\linewidth]{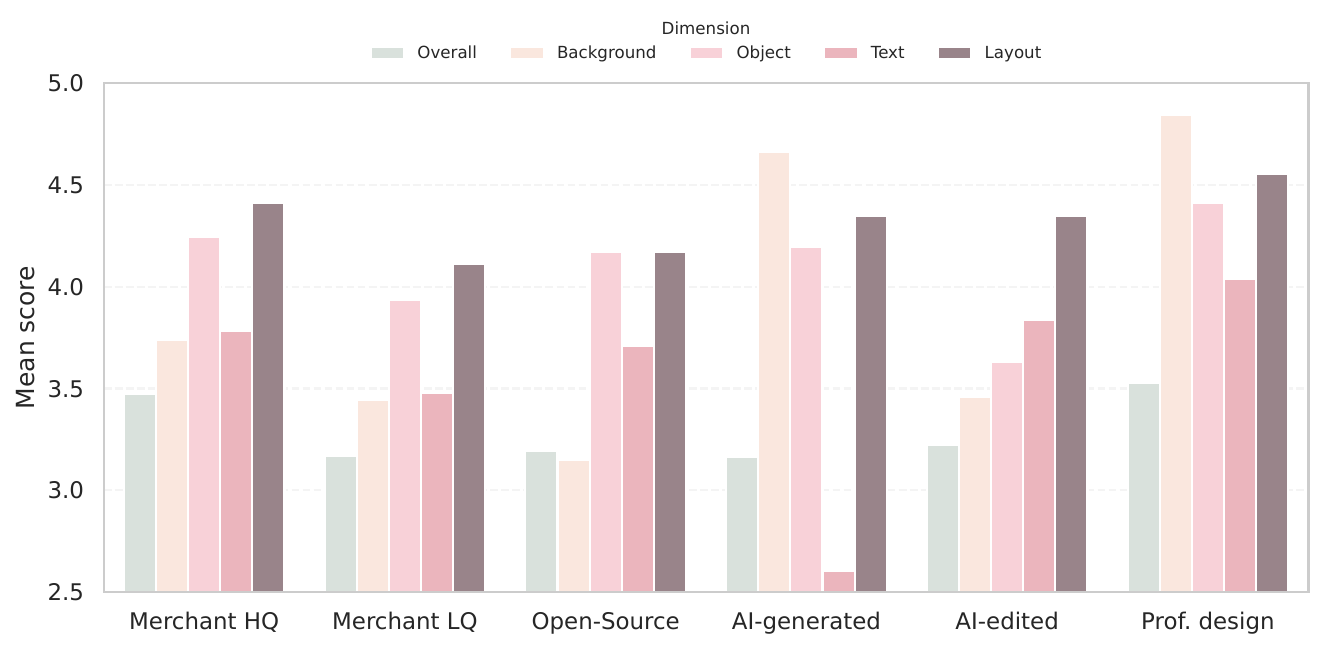}
    \caption{
        \textbf{Mean expert scores by source on E-comIQ-18k.}
    }
    \label{fig:datastastic}
\end{figure}

\begin{figure*}[t]
    \centering
    \includegraphics[width=\linewidth]{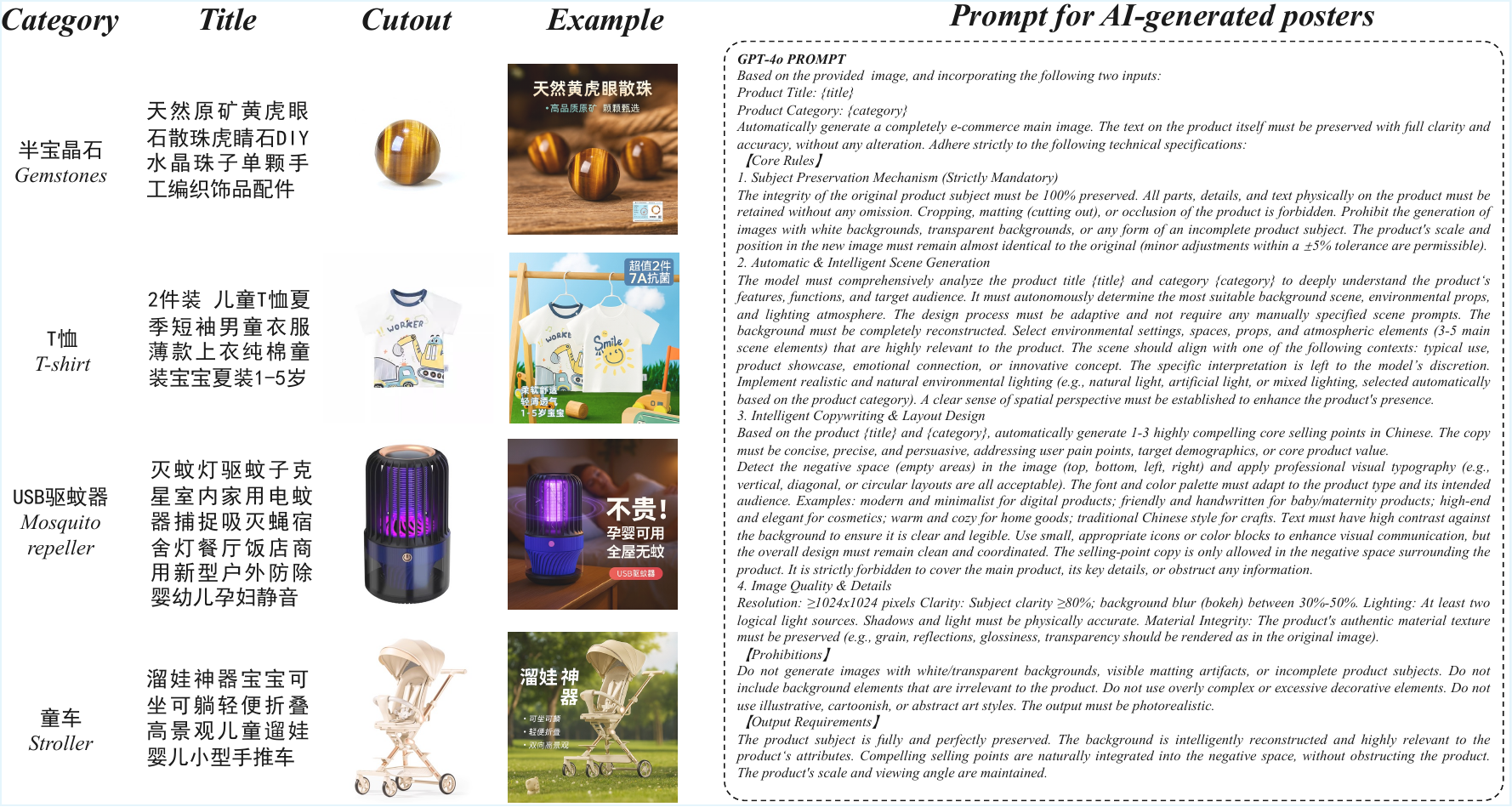}
    \caption{\textbf{Examples and prompt for the AI-generated subset.}
For each product we take the original category, Chinese title, and a white-background cutout (left),
and use GPT-4o with the prompt template on the right to generate a complete e-commerce poster (middle).
The prompt enforces strict subject preservation, automatic scene design, and Chinese selling-point copy,
so that the generated posters are photorealistic and commercially usable.
    }
    \label{fig:soracase}
\end{figure*}

For each source we compute the mean scores on Overall, Background, Object,
Text, and Layout to characterise its quality profile; the statistics are
reported in Fig~\ref{fig:datastastic}.

\begin{table}[thbp]
\centering
\footnotesize
\setlength{\tabcolsep}{17pt}

\caption{\textbf{Train/val/test splits of E-comIQ-18k.}}
\label{tab:source_split}
\begin{tabular}{lccc}
\toprule
\textbf{Source} & \textbf{Train} & \textbf{Val} & \textbf{Test} \\
\midrule
Merchant HQ    & 4166 & 555 & 279 \\
Merchant LQ    & 4166 & 555 & 279 \\
Open-Source    & 2500 & 333 & 167 \\
AI-generated   & 1666 & 222 & 112 \\
AI-edited      & 1666 & 222 & 112 \\
Prof. design   & 833  & 111 & 56  \\
\bottomrule
\end{tabular}
\end{table}

\subsection{Annotation Checklist and Tag Taxonomy}   
\label{app:checklist}

Table~\ref{tab:checklist} lists the checklist used in E-comIQ-18k. For each
image, experts annotate four dimensions (Background, Object, Text, and
Layout) with multi-label issue tags and a continuous score in $[1.0, 5.0]$
(one decimal allowed). Tags mark specific defects, while scores summarise
the perceived quality of that dimension. A key design choice is the separation between Object and Text. All textual
elements printed on the product itself (e.g., brand names and packaging
copy) are treated as part of the Object: blurry or malformed packaging text
is annotated under Object and only affects the Object score. The Text
dimension covers only overlaid marketing copy (titles, slogans, prices,
callouts, etc.), where issues such as incorrect line breaks, irrelevant or
redundant content, stroke rendering errors, missing text, overlap, or
inappropriate font size are recorded. Background and Layout tags focus on
global presentation (scene suitability, clutter, balance, occlusion), and an
Overall score summarises the commercial usability of the poster given all dimensions.

\subsection{Annotation Interface and Reliability}    

\label{app:interface_reliability}

For CoT rationales, we provide a dedicated human-AI collaboration view. 
Given the expert scores, tags, and image, Qwen-2.5-VL-Max first generates a rationale draft. 
In the interface, the image is shown on the left, and the model-generated paragraph is shown on the right. 
Annotators edit the text in a span-based NER-style manner: they can highlight spans to delete, replace them with corrected wording, or insert short additions where the explanation is incomplete; sentences that are entirely incorrect are simply struck out.
All edits are recorded, and we compute a character-level edit rate in Chinese, with an average of 32.3\% and a maximum of 83.39\%, indicating that substantial human refinement is often required.



\begin{table}[t]
    \centering
    \small
    \setlength{\tabcolsep}{4pt}
    \caption{\textbf{Annotation checklist and tag taxonomy.}}
    \label{tab:checklist}
    \begin{tabular}{p{0.92\linewidth}}
        \toprule
        \textbf{Dimension / Issue tags} \\
        \midrule
        \textit{Background} \\[2pt]
        $\square$ Color clash with product or brand;\\[2pt]
        $\square$ weak scene or context;\\[2pt]
        $\square$ irrelevant scene;\\[2pt]
        $\square$ cluttered or noisy background;\\[2pt]
        $\square$ strong ``AI-generated'' artefacts;\\[2pt]
        $\square$ missing or broken body parts;\\[2pt]
        $\square$ heavy cut-and-paste / compositing artefacts;\\[2pt]
        Other tags: \underline{\hspace{5cm}};\\[2pt]
        \textbf{Score:} \underline{\hspace{1cm}} \\  
        \midrule
        \textit{Object} \\[2pt]
        $\square$ Illegible or blurry text on the product packaging;\\[2pt]
        $\square$ incomplete object contour (parts missing or cut off);\\[2pt]
        $\square$ extra or duplicated parts (contour overgrowth);\\[2pt]
        $\square$ physically implausible placement or pose;\\[2pt]
        $\square$ lighting or perspective inconsistent with the scene;\\[2pt]
        $\square$ unreasonable scale or proportion;\\[2pt]
        $\square$ visible compositing artefacts;\\[2pt]
        Other tags: \underline{\hspace{5cm}};\\[2pt]
        \textbf{Score:} \underline{\hspace{1cm}}  \\
        \midrule
        \textit{Text} \\[2pt]
        $\square$ Incorrect or awkward line breaks;\\[2pt]
        $\square$ content irrelevant to the product or promotion;\\[2pt]
        $\square$ style mismatch with brand or poster tone;\\[2pt]
        $\square$ stroke rendering errors;\\[2pt]
        $\square$ spelling mistakes or typos;\\[2pt]
        $\square$ missing expected overlaid text;\\[2pt]
        $\square$ font too large;\\[2pt]
        $\square$ font too small;\\[2pt]
        $\square$ overlapping text (with other text or the object);\\[2pt]
        $\square$ redundant or repetitive text;\\[2pt]
        Other tags: \underline{\hspace{5cm}};\\[2pt]
        \textbf{Score:} \underline{\hspace{1cm}} \\
        \midrule
        \textit{Layout} \\[2pt]
        $\square$ Overly crowded or cluttered layout;\\[2pt]
        $\square$ excessive empty space;\\[2pt]
        $\square$ visually unbalanced composition;\\[2pt]
        $\square$ important elements occluded or mutually blocking;\\[2pt]
        Other tags: \underline{\hspace{5cm}};\\[2pt]
        \textbf{Score:} \underline{\hspace{1cm}} \\
        \midrule
        \textit{Overall} \\[2pt]
        \textbf{Score:} \underline{\hspace{1cm}}\\
        \bottomrule
    \end{tabular}
\end{table}

\begin{figure*}[t]
    \centering
    \includegraphics[width=\linewidth]{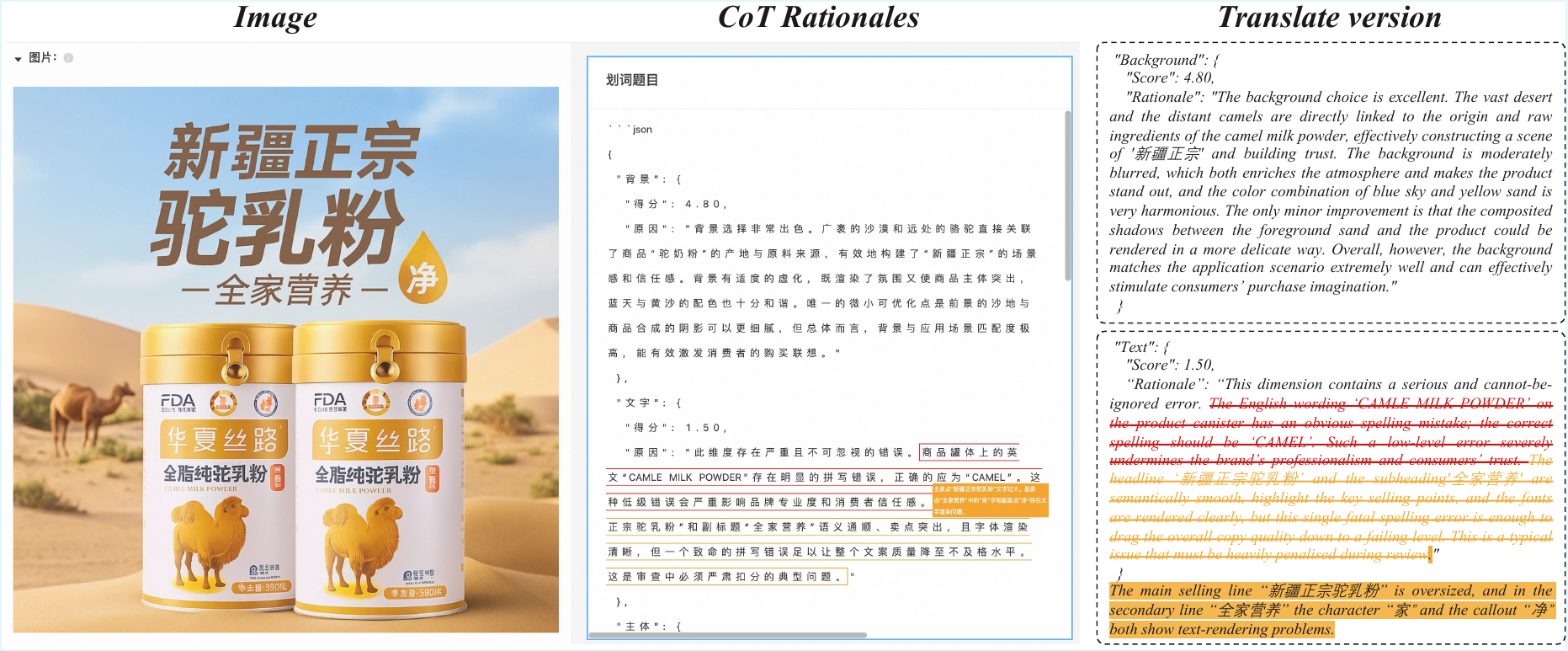}
    \caption{\textbf{Example of our CoT editing interface.}
    Given a poster image (left), the annotator reviews the LLM-generated Chinese rationale (middle) and performs span-level edits to correct errors (shown in red/orange), producing an English translated version (right) that remains faithful to expert judgement.}
    \label{fig:nercase}
\end{figure*}
\begin{figure*}[t]
    \centering
    \begin{minipage}{0.48\textwidth}
        \centering
        \includegraphics[width=\linewidth]{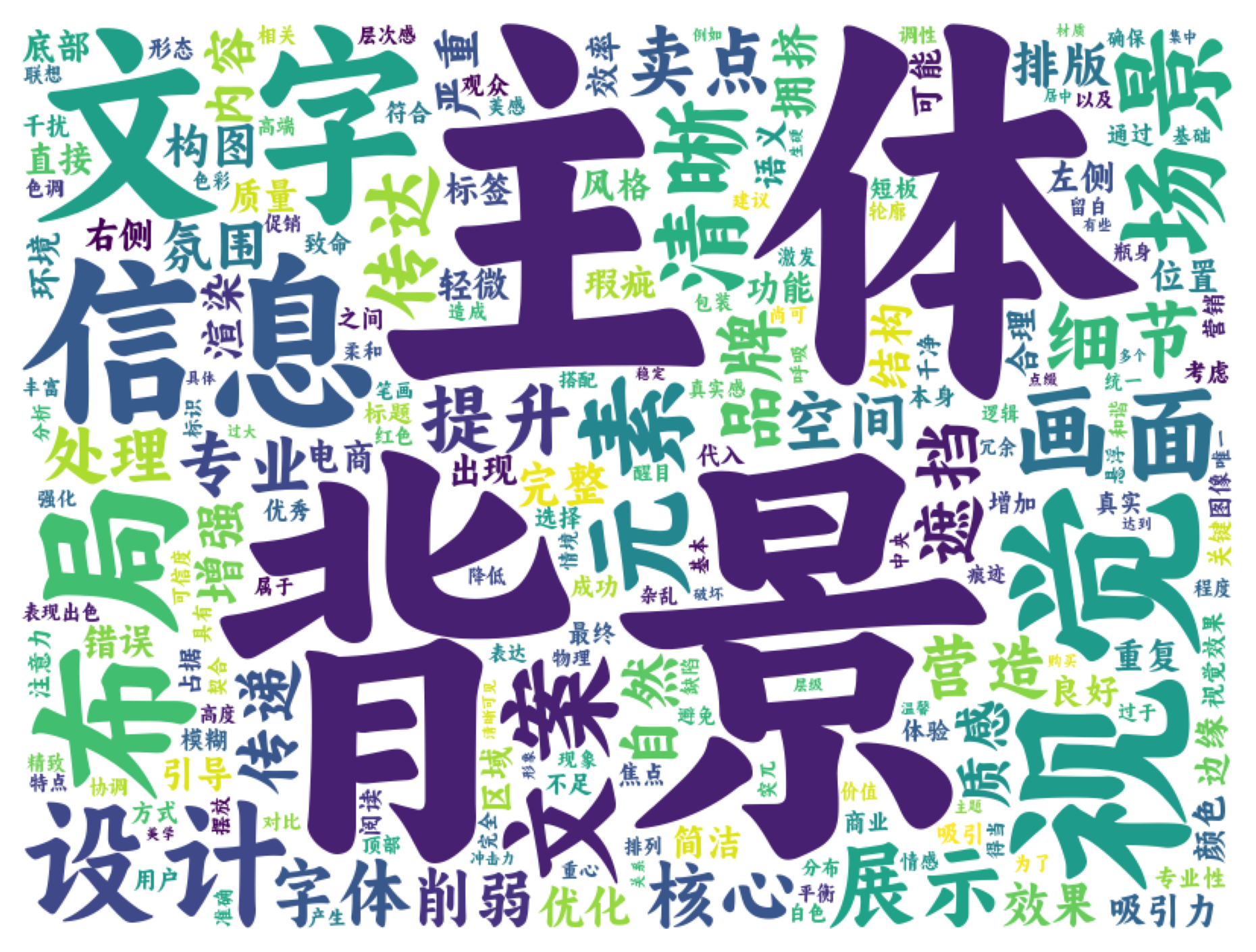}
        \caption*{\textbf{(a) Chinese CoT Word Cloud}}
    \end{minipage}
    \hfill
    \begin{minipage}{0.48\textwidth}
        \centering
        \includegraphics[width=\linewidth]{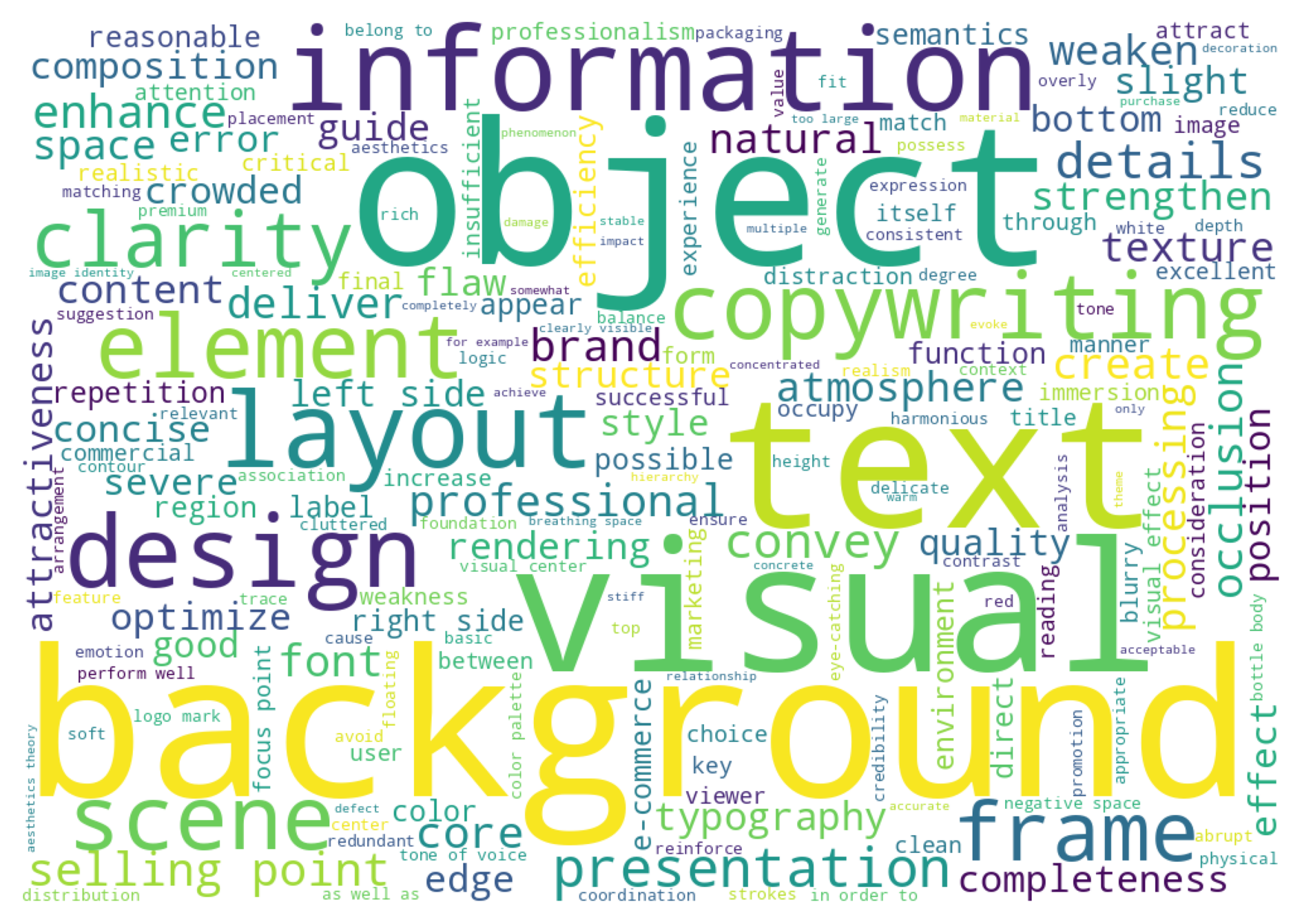}
        \caption*{\textbf{(b) English CoT Word Cloud}}
    \end{minipage}

    \caption{
        \textbf{Word frequency analysis of model reasoning traces.}
        We visualize the top frequent words appearing in the model’s chain-of-thought (CoT) across 18k e-commerce samples. 
        The Chinese word cloud (left) is directly computed from the original CoT supervision, while the English version (right) is obtained by carefully mapping the top 200 Chinese terms to semantically aligned English phrases. 
        Both visualizations reflect consistent semantic emphasis on background, object clarity, textual quality, composition, and visual communication.
    }
    \label{fig:cot_wordcloud}
\end{figure*}


\renewcommand{\thesection}{B}

\section{E-comIQ-M: Model and Training Details}
\subsection{Model Configuration}                   
\label{app:model_config}

We build E-comIQ-M on the Qwen2.5-VL-7B-Instruct, using the official vision encoder and tokenizer without architectural changes. 

Each sample contains a single poster image and an evaluation instruction. The images are decoded as RGB and resized to $512\times512$ before being fed into the model. Both SFT and GRPO use the same instruction-following format. Given the instruction, the model first produces a Chain-of-Thought in natural-language inside a \texttt{<think></think>} block and then outputs a JSON object inside a \texttt{<answer></answer>} block.

During training and evaluation, any sample whose \texttt{<answer></answer>} block cannot be parsed as valid JSON is treated as invalid , and at inference time we retry decoding up to three times ( $temperature=1.0$,  $top\_p=0.95$, maximum 4096 new tokens, single-turn generation).

\begin{figure}[htbp]
    \centering
    \includegraphics[width=\linewidth]{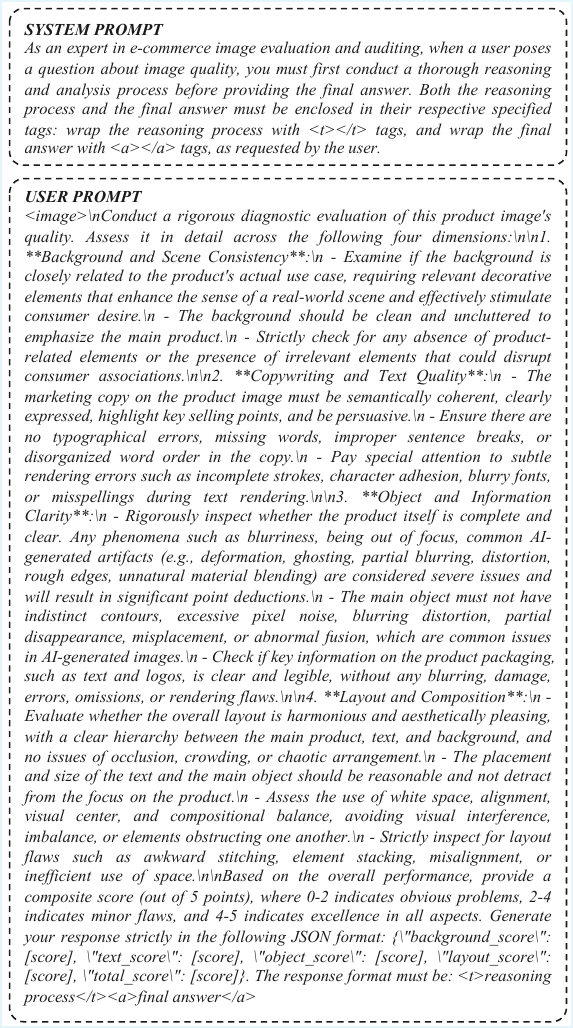}
\caption{\textbf{Instruction and prompt template for E-comIQ-M.}
We use a fixed system prompt and a fixed user prompt that ask the model to first provide a \texttt{<think>} Chain-of-Thought and then output a JSON object with five scores in the \texttt{<answer>} block.}
    \label{fig:prompt}
\end{figure}
\subsection{SFT Training Hyperparameters}           
\label{app:sft}

We perform supervised fine-tuning on the 15k training images using the
instruction format in Fig.~\ref{fig:prompt}. We use AdamW with a learning rate of
$5\times10^{-5}$ and a cosine schedule without warm-up. All parameters of
Qwen-2.5-VL-7B-Instruct are updated (full fine-tuning) within the
LLaMA-Factory framework.
SFT is run with batch size per-GPU $=1$ and gradient
accumulation $=8$, giving 64 samples per optimizer update. We train for
3 epochs using DeepSpeed ZeRO stage~3 with bf16 precision and a
maximum sequence length of 4096 tokens; when the CoT exceeds this limit, only
the tail of the reasoning is truncated, while the JSON answer is always kept
intact.
\subsection{GRPO Training Details}                  
\label{app:grpo}

For the second stage we apply GRPO on a hard subset $\mathcal{D}_{\text{hard}}$
of 3k training samples. We initialise the policy $\pi_\theta$ from the SFT
checkpoint and use the same model as the frozen reference $\pi_{\text{ref}}$.
The instruction format and target JSON schema are identical to SFT
(Fig.~\ref{fig:prompt}). For each prompt we sample a group of
$G=4$ continuations with  $temperature=1.0$,  $top\_p=0.95$ and $max\_new\_tokens=4096$. Invalid JSON outputs are re-sampled up to
three times; if all attempts fail, the sample is marked invalid and its
reward is set to zero.

Training is carried out on the 3k hard samples for 500 optimizer steps. At
each step we process one prompt per device with group size $G=4$ and use
gradient accumulation $=8$, which yields 32 trajectories per update. We use
AdamW with a cosine learning rate schedule, base learning rate
$1\times10^{-6}$ and warmup ratio $0.1$. The KL regularisation coefficient
$\beta$ in GRPO is set to $0.1$. As shown in our logs, the training remains
stable without reward collapse.
\vspace{-10pt}
\paragraph{Hard subset construction.}
\begin{table}[t]
\centering
\small
\setlength{\tabcolsep}{6pt}
\caption{\textbf{Effect of hard-subset size on GRPO performance.} 
Performance is reported as PLCC and SRCC on the validation split. 
The default configuration (3k) is highlighted.}
\label{tab:hard_size}
\begin{tabular}{lccccc}
\toprule
\textbf{Metric} & \textbf{1k} & \textbf{2k} & \textbf{3k (default)} & \textbf{4k} & \textbf{5k} \\
\midrule
PLCC & 0.412 & 0.421 & 0.425 & 0.426 & 0.423 \\
SRCC & 0.423 & 0.429 & 0.433 & 0.431 & 0.430 \\
\bottomrule
\end{tabular}
\end{table}

We select $\mathcal{D}_{\text{hard}}$ from the 15k SFT training set using the
SFT model's regression error while preserving the source distribution.
For each sample $j$ we denote the expert scores by
$\mathbf{y}_j \in \mathbb{R}^5$ and the SFT prediction by
$\hat{\mathbf{y}}_j \in \mathbb{R}^5$ (four dimensions plus overall).
We first compute the mean squared error 
$\mathrm{MSE}_j = \|\mathbf{y}_j - \hat{\mathbf{y}}_j\|_2^2$, 
then rank samples within each source by $\mathrm{MSE}_j$ and take the top
fraction so that the final hard subset contains 3k examples with a source
mix matching the original 15k set.
To verify that our results are not overly sensitive to this choice, we also
vary the hard-subset size from 1k to 5k and re-train GRPO. As shown in
Table~\ref{tab:hard_size}, overall PLCC and SRCC on the validation set remain
stable across different sizes, with the 3k configuration achieving a slightly
better balance between performance and computational cost. We therefore use
3k as the default setting in all main experiments.



\begin{algorithm}
\footnotesize
    \renewcommand{\algorithmicrequire}{\textbf{Input:}}
    \renewcommand{\algorithmicensure}{\textbf{Output:}}
    \caption{Construction of hard subset $\mathcal{D}_{\text{hard}}$ for GRPO}
    \label{alg:hard_subset}
    \begin{algorithmic}[1]
        \REQUIRE Training set $\mathcal{D}_{\text{train}}=\{(x_j,\mathbf{y}_j,s_j)\}_{j=1}^{N}$, \\
                 \quad SFT model $f_{\text{SFT}}$, \\
                 \quad target size $K$ (here $K=3000$)
        \ENSURE Hard subset $\mathcal{D}_{\text{hard}}$
        \STATE \textbf{Constants:}
        \STATE \quad Number of dimensions $D=5$ (4 sub-scores + overall)
        \STATE \quad Source set $\mathcal{S}=\{\text{HQ},\text{LQ},\text{Open-Source},\text{AI-gen.},\text{AI-edit},\text{Prof.}\}$
        \STATE Initialize per-source container $\mathcal{B}_s \gets [\;]$ for all $s \in \mathcal{S}$
        \FOR{$j = 1$ to $N$}
            \STATE $\hat{\mathbf{y}}_j \gets f_{\text{SFT}}(x_j)$ \COMMENT{SFT prediction}
            \STATE $e_j \gets \frac{1}{D}\|\hat{\mathbf{y}}_j - \mathbf{y}_j\|_2^2$ \COMMENT{mean squared error}
            \STATE Append $(x_j,\mathbf{y}_j,e_j)$ to $\mathcal{B}_{s_j}$
        \ENDFOR
        \STATE Compute per-source sizes $N_s \gets |\mathcal{B}_s|$ for all $s \in \mathcal{S}$
        \STATE Set $K_s \gets \left\lfloor K \cdot \frac{N_s}{\sum_{s' \in \mathcal{S}} N_{s'}} \right\rfloor$ for all $s \in \mathcal{S}$
        \STATE Initialize $\mathcal{D}_{\text{hard}} \gets \emptyset$
        \FOR{each source $s \in \mathcal{S}$}
            \STATE Sort $\mathcal{B}_s$ in \textbf{descending} order of $e_j$
            \STATE Take the first $K_s$ samples from $\mathcal{B}_s$ and add them to $\mathcal{D}_{\text{hard}}$
        \ENDFOR
        \STATE \textbf{return} $\mathcal{D}_{\text{hard}}$
    \end{algorithmic}
\end{algorithm}

\subsection{Reward Design and Ablations}
\label{app:reward_ablation}

As discussed in Sec. 4.2 and Table 5,
combining the accuracy and distribution terms ($R_{\text{acc}} + R_{\text{dist}}$) on
top of SFT gives the best overall performance. Here we provide additional
analysis on the reward weights and GRPO optimisation hyperparameters.

\definecolor{thr05}{HTML}{d4e09b}
\definecolor{thr04}{HTML}{f6f4d2}
\definecolor{thr03}{HTML}{cbdfbd}
\definecolor{thr02}{HTML}{a44a3f}
\definecolor{thr01}{HTML}{f19c79}

\begin{figure*}[htbp]
    \centering
    \includegraphics[width=0.47\textwidth]{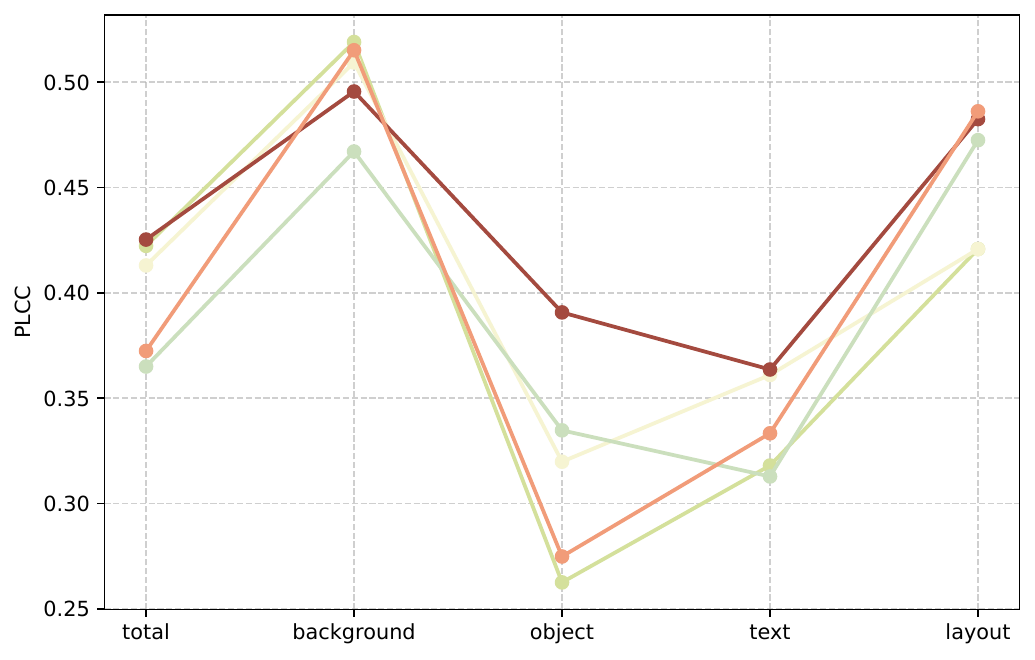}
    \hfill
    \includegraphics[width=0.47\textwidth]{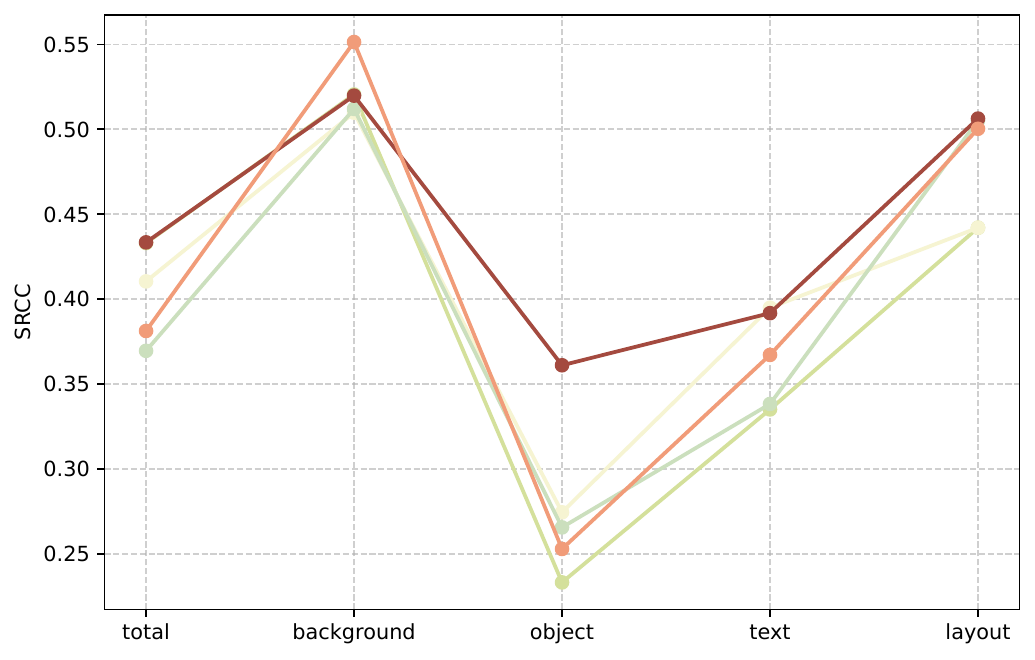}

    \caption*{
        \parbox[c][8pt][l]{10pt}{\colorbox{thr05}{}}\,$\tau=0.5$\quad
        \parbox[c][8pt][l]{10pt}{\colorbox{thr04}{}}\,$\tau=0.4$\quad
        \parbox[c][8pt][l]{10pt}{\colorbox{thr03}{}}\,$\tau=0.3$\quad
        \parbox[c][8pt][l]{10pt}{\colorbox{thr02}{}}\,$\tau=0.2$\quad
        \parbox[c][8pt][l]{10pt}{\colorbox{thr01}{}}\,$\tau=0.1$
    }

    \caption{
        \textbf{Effect of the accuracy tolerance $\tau$ on reward performance.}
        Left: overall PLCC across the four sub-dimensions (Background, Object, Text, Layout) and the total score under different $\tau$ values.
        Right: corresponding SRCC results.
        Each coloured line denotes a different tolerance setting.
    }
    \label{fig:threshold_ablation}
\end{figure*}

\definecolor{w100}{HTML}{d4e09b} 
\definecolor{w085}{HTML}{f6f4d2} 
\definecolor{w075}{HTML}{cbdfbd} 
\definecolor{w065}{HTML}{a44a3f} 
\definecolor{w055}{HTML}{e6beae} 
\definecolor{w045}{HTML}{bee9e8} 
\definecolor{w035}{HTML}{c6ac8f} 

\begin{figure*}[htbp]
    \centering
    \includegraphics[width=0.47\textwidth]{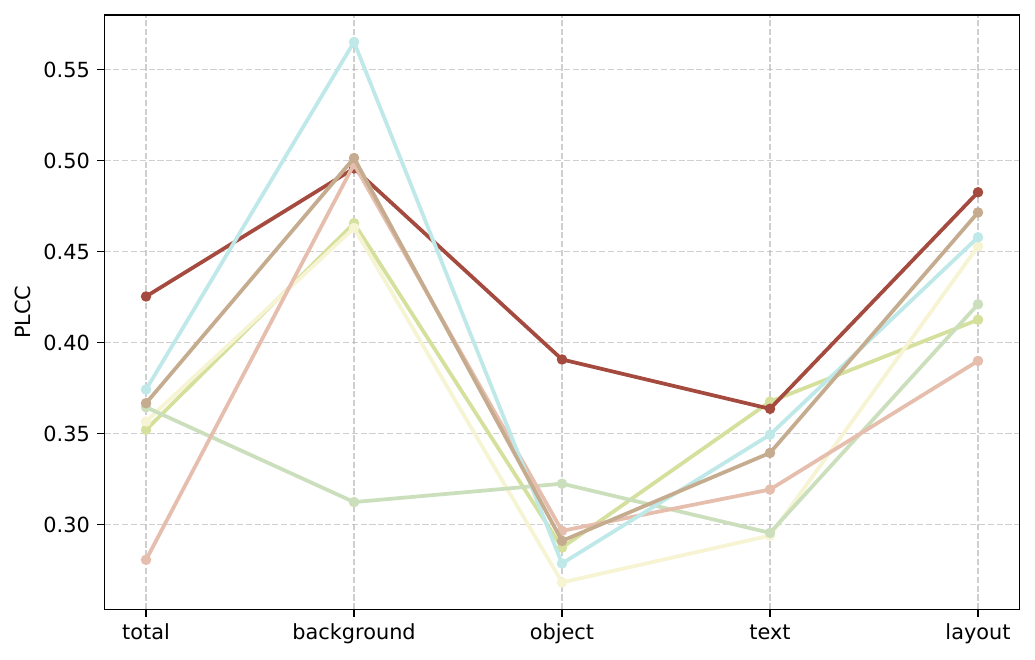}
    \hfill
    \includegraphics[width=0.47\textwidth]{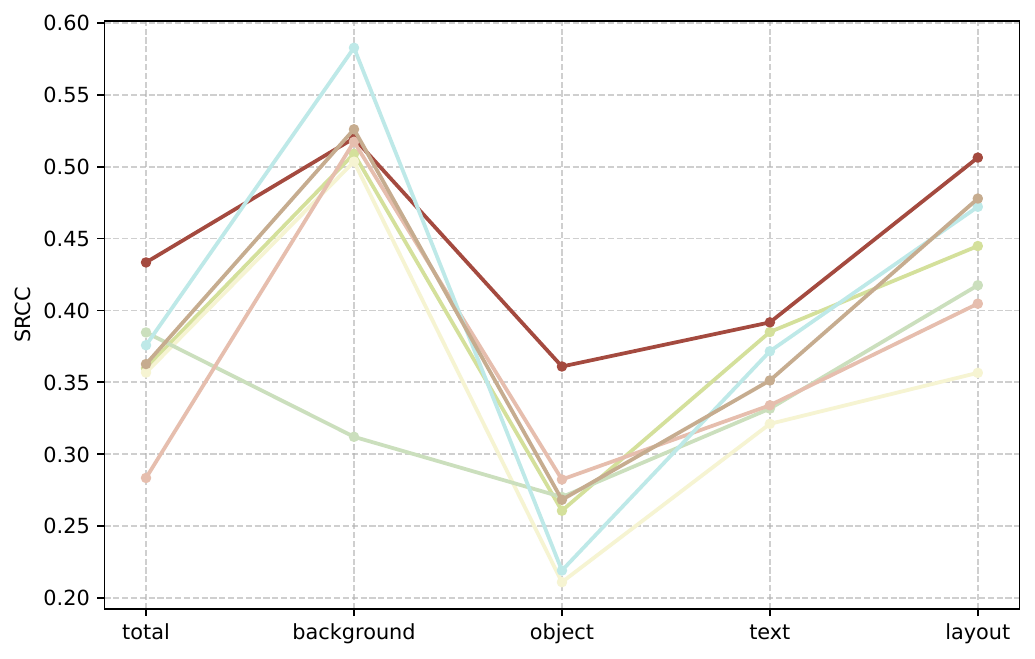}

    \caption*{
        \parbox[c][8pt][l]{10pt}{\colorbox{w100}{}}\,$\lambda_{\text{score}}=1.00$\quad
        \parbox[c][8pt][l]{10pt}{\colorbox{w085}{}}\,$\lambda_{\text{score}}=0.85$\quad
        \parbox[c][8pt][l]{10pt}{\colorbox{w075}{}}\,$\lambda_{\text{score}}=0.75$\quad
        \parbox[c][8pt][l]{10pt}{\colorbox{w065}{}}\,$\lambda_{\text{score}}=0.65$\quad\\[2pt]
        \parbox[c][8pt][l]{10pt}{\colorbox{w055}{}}\,$\lambda_{\text{score}}=0.55$\quad
        \parbox[c][8pt][l]{10pt}{\colorbox{w045}{}}\,$\lambda_{\text{score}}=0.45$\quad
        \parbox[c][8pt][l]{10pt}{\colorbox{w035}{}}\,$\lambda_{\text{score}}=0.35$
    }

    \caption{
        \textbf{Effect of the accuracy weight $\lambda_{\text{score}}$ on reward performance.}
        Left: PLCC across the four sub-dimensions (Background, Object, Text, Layout) and the total score under different $\lambda_{\text{score}}$ values.
        Right: corresponding SRCC results.
        Each coloured line denotes a different setting of the accuracy weight in the reward.
    }
    \vspace{-15pt}
    \label{fig:lambda_ablation}
\end{figure*}

\vspace{-15pt}
\paragraph{Reward weight sensitivity.}
Figure~\ref{fig:threshold_ablation} studies the effect of the accuracy
tolerance $\tau$ in $R_{\text{acc}}$. Very loose thresholds
($\tau\geq 0.5$) make the reward less informative and lead to weaker PLCC
and SRCC, while very strict thresholds ($\tau=0.1$) also hurt performance.
The curves are most stable around $\tau=0.2$, which we adopt as the default.
Figure~\ref{fig:lambda_ablation} varies the accuracy weight
$\lambda_{\text{score}}$ that balances $R_{\text{acc}}$ and $R_{\text{dist}}$.
Using only the accuracy term ($\lambda_{\text{score}}=1.0$) or giving it too
little weight ($\lambda_{\text{score}}\le 0.45$) degrades both correlations.
The best trade off is obtained near $\lambda_{\text{score}}=0.65$, confirming
that a moderate contribution from the distribution term helps align the
geometry of sub scores with expert ratings.

\vspace{-15pt}
\paragraph{GRPO optimisation hyperparameters.}
We further sweep the GRPO learning rate and KL penalty coefficient $\beta$
(see Table~\ref{tab:grpo_hparam}). Very small learning rates or $\beta$
values slow down optimisation and yield limited gains over SFT, while larger
values lead to unstable training and a drop in PLCC/SRCC. The final setting
used in the main experiments ($\text{lr}=1\times10^{-6}$, $\beta=0.1$) lies
in a stable region and offers the best overall balance between convergence
speed and evaluation performance.
\begin{table*}[t]
\centering
\caption{
\textbf{Effect of GRPO learning rate and KL coefficient on correlation performance.}
Each cell reports \textbf{PLCC / SRCC} on the E-comIQ-18k test set.
}
\label{tab:grpo_hparam}
\footnotesize
\setlength{\tabcolsep}{10pt}
\begin{tabular}{cc|cc|cc|cc|cc|cc}
\toprule
\multicolumn{2}{c|}{\textbf{Setting}} &
\multicolumn{2}{c|}{\textbf{Overall}} &
\multicolumn{2}{c|}{\textbf{Background}} &
\multicolumn{2}{c|}{\textbf{Object}} &
\multicolumn{2}{c|}{\textbf{Text}} &
\multicolumn{2}{c}{\textbf{Layout}} \\
$\textit{lr}$ & $\boldsymbol{\beta}$ &
\textit{PLCC} & \textit{SRCC} &
\textit{PLCC} & \textit{SRCC} &
\textit{PLCC} & \textit{SRCC} &
\textit{PLCC} & \textit{SRCC} &
\textit{PLCC} & \textit{SRCC} \\
\midrule
$1\times10^{-6}$ & 0.10 &
\textbf{0.425} & \textbf{0.433} &
\textbf{0.496} & \underline{0.520} &
\textbf{0.391} & \textbf{0.361} &
\textbf{0.364} & \textbf{0.392} &
\textbf{0.483} & \textbf{0.506} \\
$5\times10^{-5}$ & 0.10 &
0.267 & 0.277 &
0.383 & 0.462 &
0.189 & 0.223 &
0.270 & 0.270 &
0.375 & 0.403 \\
$1\times10^{-6}$ & 0.05 &
0.329 & \underline{0.348} &
0.441 & 0.508 &
0.263 & \underline{0.239} &
\underline{0.342} & \underline{0.346} &
0.431 & 0.450 \\
$5\times10^{-5}$ & 0.05 &
0.326 & 0.302 &
0.417 & 0.454 &
\underline{0.265} & 0.237 &
0.226 & 0.311 &
\underline{0.455} & \underline{0.488} \\
$1\times10^{-6}$ & 0 &
\underline{0.346} & 0.346 &
\underline{0.458} & \textbf{0.530} &
0.272 & 0.238 &
0.272 & 0.283 &
0.390 & 0.418 \\
$5\times10^{-5}$ & 0 &
0.265 & 0.235 &
0.312 & 0.312 &
0.123 & 0.070 &
0.096 & 0.132 &
0.221 & 0.218 \\
\bottomrule
\end{tabular}
\end{table*}

\begin{figure*}[htbp]
    \centering
    \begin{minipage}[b]{0.48\linewidth}
        \centering
        \includegraphics[width=\linewidth]{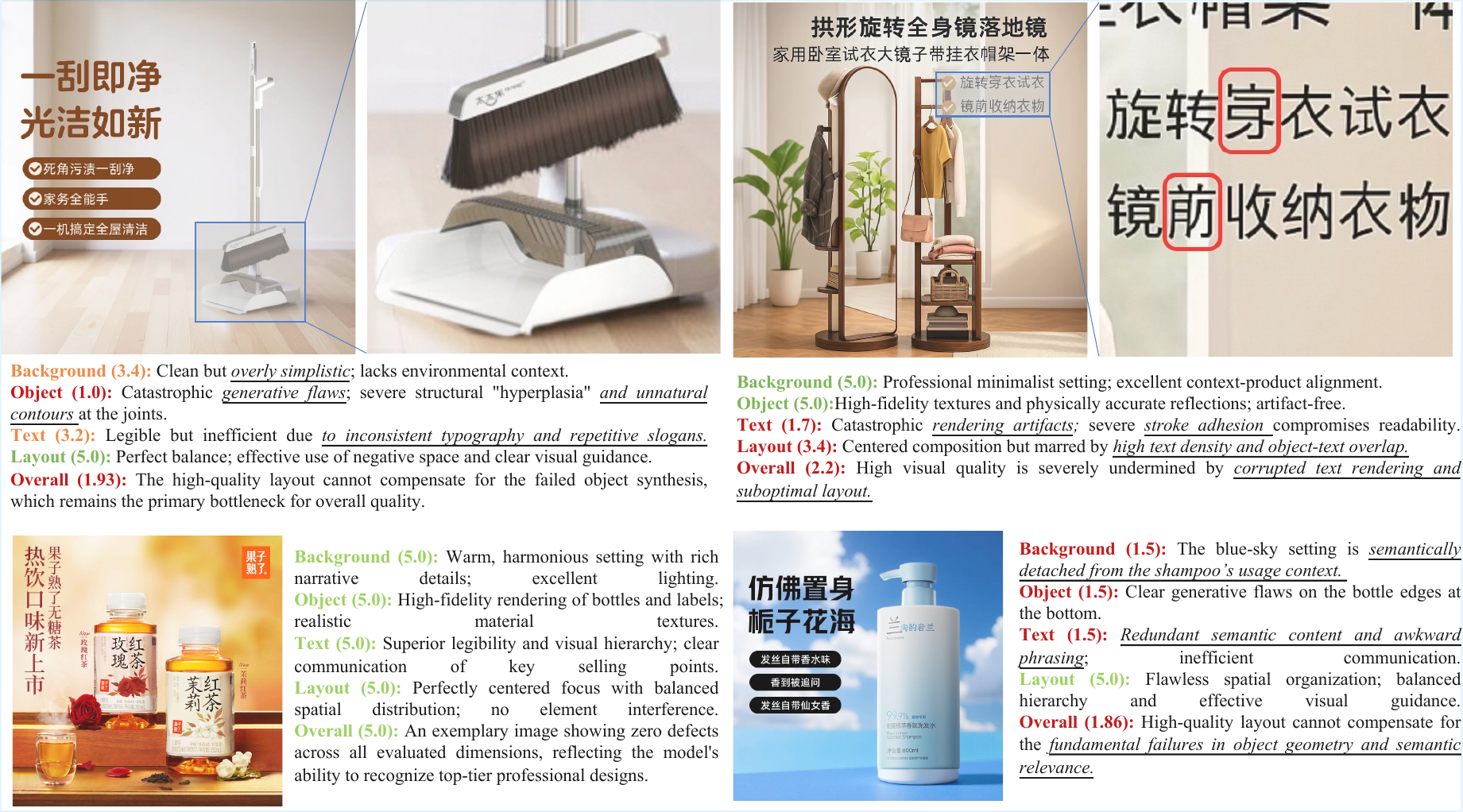}
    \end{minipage}%
    \hfill
    \begin{minipage}[b]{0.48\linewidth}
        \centering
        \includegraphics[width=\linewidth]{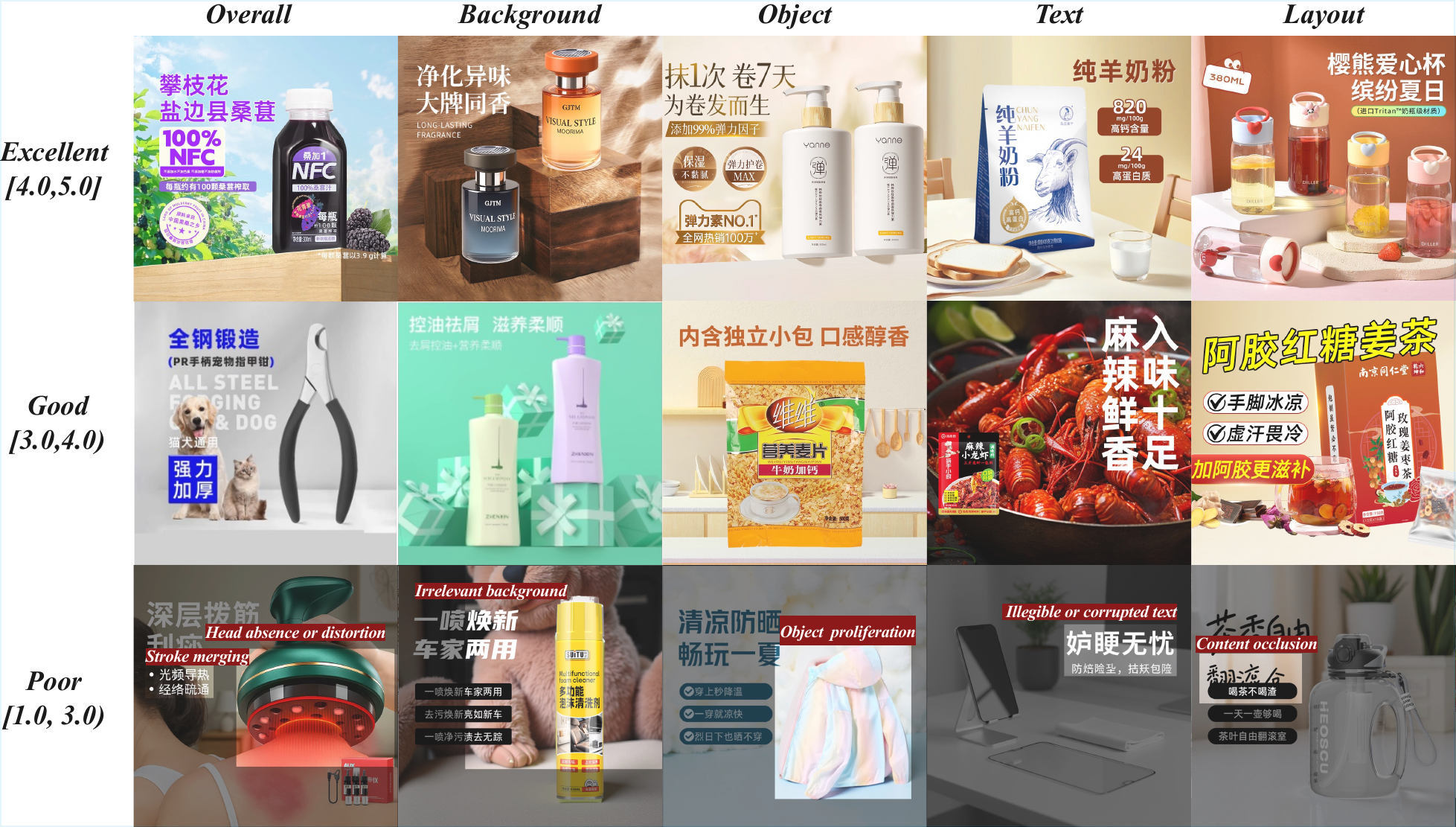}
    \end{minipage}
    \caption{Additional qualitative examples.}
    \label{fig:casestudy}
\end{figure*}


\renewcommand{\thesection}{C}

\section{E-comIQ-Bench and Evaluation Toolbox}

\subsection{Prompt Design and Generation Setup}     
\label{app:prompt_and_generation}

The construction procedure of E-comIQ-Bench follows the design described in
Sec. 5.1. Each case contains a foreground cutout,

its merchant poster, and a Chinese prompt derived from the product’s selling
points. Here we provide additional implementation details regarding prompt
generation and image synthesis.
\begin{figure*}[t]
    \centering
    \includegraphics[width=\linewidth]{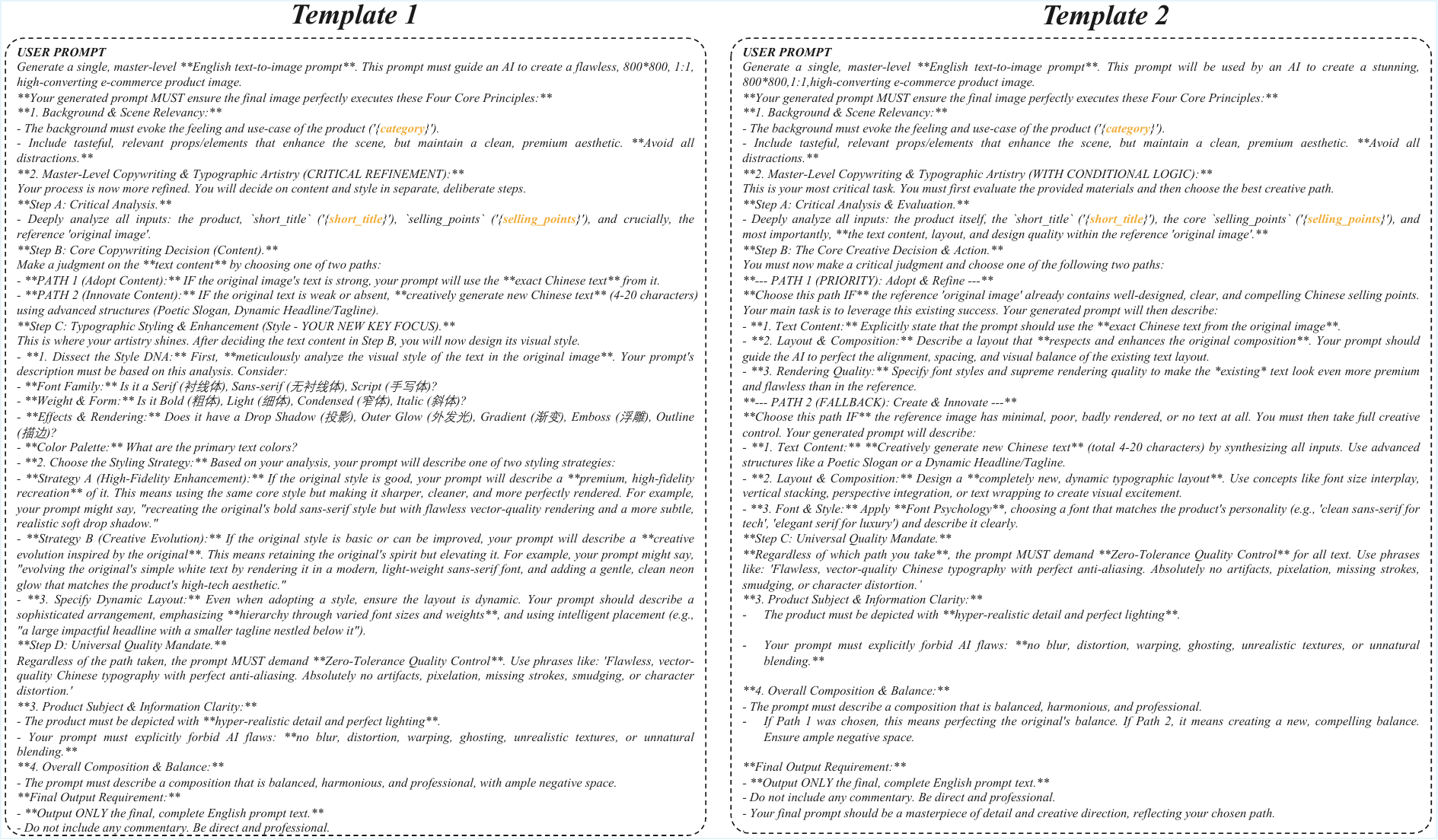}
    \caption{
    \textbf{Prompt templates for generating Chinese e-commerce poster instructions.}
    We provide the white-background cutout, the original merchant poster, and structured attributes (category, short title, and selling points) to Qwen2.5-VL-72B.
    The model rewrites the information into a professional, stylistic \emph{generation prompt} that follows detailed typography, layout, and text-quality constraints.
    Five templates are used to encourage stylistic diversity.
    }
    \label{fig:promptemplate}
\end{figure*}
\begin{figure*}[t]
    \centering
    \includegraphics[width=\linewidth]{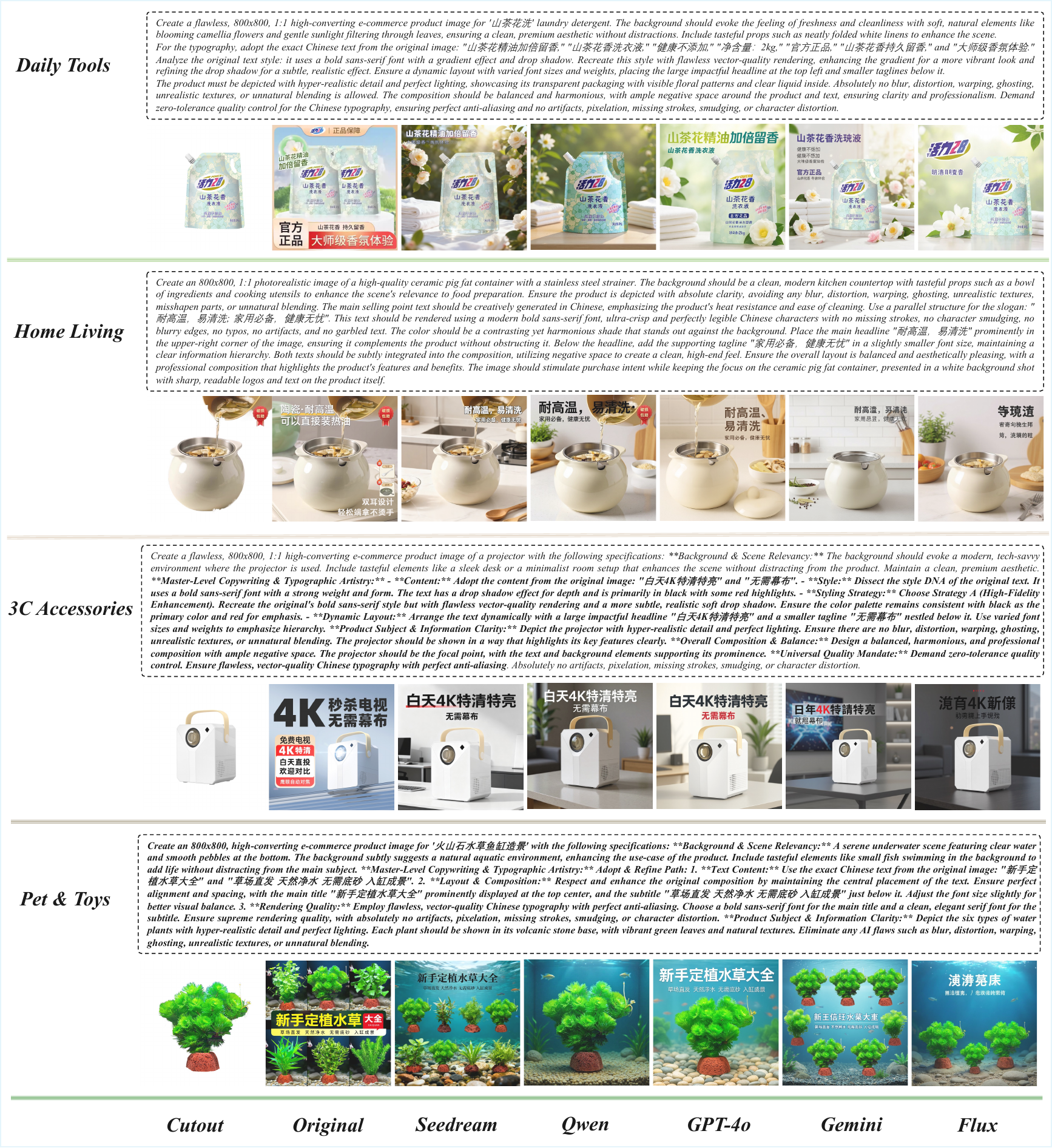}
    \caption{
    \textbf{Prompt–generation showcase across product categories (1/2).}
    For each cutout–prompt pair, several commercial systems and research models are queried to generate one poster per model.
    The merchant poster is shown as the human-designed reference.
    Categories on this page include \emph{Daily Tools, Home Living}, and \emph{3C Accessories}.
    }
    \label{fig:benchmark-showcase_1}
\end{figure*}
\begin{figure*}[t]
    \centering
    \includegraphics[width=\linewidth]{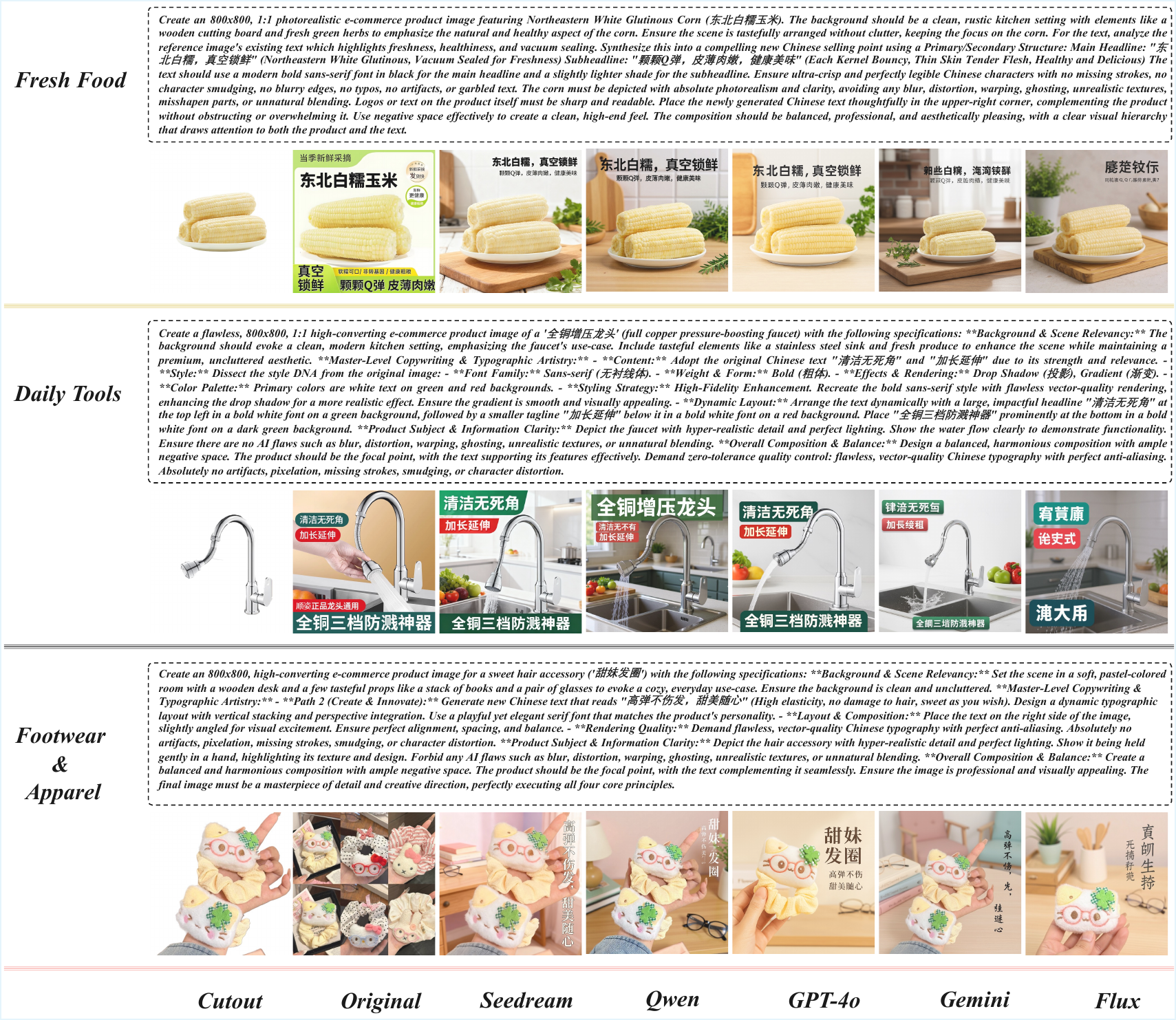}
    \caption{
    \textbf{Prompt–generation showcase across product categories (2/2).}
    The same protocol applies to the remaining categories: \emph{Fresh Food, Daily Tools (additional cases), Footwear \& Apparel}, and \emph{Pet \& Toys}.
    }
    \label{fig:benchmark-showcase_2}
\end{figure*}
\vspace{-10pt}
\paragraph{Prompt generation.}
Building on Sec. 5.1, we now detail how the Chinese
poster prompts are constructed. Given a product cutout, the original merchant
poster, and structured product attributes (category, short title, and selling
points), we first query Qwen2.5-VL-72B to generate a high-level poster prompt.
To reduce prompt bias, we design five template variants covering different
copywriting styles, typography rules, and layout strategies. One template is
randomly selected and filled with the extracted attributes, and the model
rewrites the text with improved commercial tone and layout instructions. The
output is a complete ``generation prompt’’ that will later be used to produce
the final poster image.

Importantly, the template itself is \emph{not} used for generation: it only
guides the rewritings made by Qwen2.5-VL-72B. The rewritten result becomes the
actual prompt supplied to different text-to-image systems (Fig.~\ref{fig:promptemplate}).
Examples from seven categories are shown in Fig.~\ref{fig:benchmark-showcase_1}
and Fig.~\ref{fig:benchmark-showcase_2}.
\vspace{-10pt}
{

\small
\setlength{\intextsep}{0pt}    
\setlength{\floatsep}{0pt}     
\setlength{\textfloatsep}{0pt} 
\setlength{\belowcaptionskip}{0pt} 

\begin{table}[b]
\centering
\small
\caption{
     Inter-annotator agreement for E-comIQ-Bench
}
\vspace{-10pt}
\label{tab:rebutal_iaa_1}
\setlength{\tabcolsep}{7pt} 
\begin{tabular}{lccccc}
\toprule
 & \textbf{Overall} & \textbf{Object} & \textbf{Background} & \textbf{Text} & \textbf{Layout} \\
\midrule
\textbf{$\alpha$} & 0.741 & 0.780 & 0.699 & 0.818 & 0.930 \\
\bottomrule
\end{tabular}
\vspace{-10pt}

\end{table}
}

\paragraph{Generation setup.}
All text-to-image models are queried with a unified image resolution of
$800\times800$ pixels, and all models support both Chinese and English prompts without requiring language-specific templates. 
We evaluate a mixture of commercial and open-source systems, including 
Seedream~4.0~\cite{seedream2025seedream}, 
GPT-4o~\cite{gpt-4o}, 
Gemini-2.5-Flash-Image~\cite{Gemini2_5_Flash_Image}, 
Flux-Kontext-max~\cite{batifol2025flux}, 
and the open-source Qwen-Image-Edit~\cite{wu2025qwen}.  
Commercial models are accessed through official HTTP APIs, whereas 
Qwen-Image-Edit is executed locally via the official SDK. 
For all models, we follow their default inference settings (e.g., sampling steps and classifier-free guidance), because our benchmark focuses on cross-model stylistic controllability rather than model-specific tuning.
A unified robustness policy is used across systems: if one query fails due to network or decoding errors, we automatically retry the same request up to three times. 
Only successful generations are kept as final benchmark samples.


\subsection{Automatic Evaluation Toolbox}           
\label{sec:toolbox_details}

E-comIQ-Bench evaluates each generated poster along the same four quality dimensions as human annotation (Background, Object, Text, Layout) plus the overall score. 
Human evaluation and our proposed E-comIQ-M serve as the two reference metrics.
In addition, two auxiliary indicators are used to measure structural correctness: object consistency and text accuracy. 

\noindent\textbf{Object consistency.}
Given a cutout, we measure how well the generated poster preserves the true object identity.
We first use DINOv2~\cite{oquab2023dinov2} to detect the target category in the poster and obtain the corresponding bounding region, then apply SAM-HQ~\cite{ke2023segment} followed by Vitmatte~\cite{yao2024vitmatte} to obtain a refined object mask. 
The extracted object region is compared against the original cutout using DINO feature cosine similarity, CLIP image embedding cosine similarity, and LPIPS perceptual distance. 
These metrics quantify semantic consistency (DINO/CLIP) and pixel-level fidelity (LPIPS) between the generated image and the ground-truth product.

\noindent\textbf{Text accuracy.}
Unlike object text on the product packaging (assessed in the Object dimension), this metric evaluates whether the generated marketing copy faithfully reflects the intended prompt semantics.
We train a lightweight text extractor to obtain structured selling-point keywords from each poster, and compare them to the prompt using two levels of textual matching:
(1) Sentence-level structural accuracy measured by F1 over detected key phrases (Phrase F1).
(2) Character-level normalised Levenshtein similarity computed as the Bag-of-Characters cosine similarity (Char Sim), which ignores ordering but enforces semantic token agreement. 
This combination penalises both missing key information and hallucinated claims.

\noindent\textbf{Reproducibility.}
All metrics are implemented in our evaluation toolbox, which will be released together with the dataset.
To ensure fairness and stability, the toolbox retries API failures up to three times and outputs merged JSON statistics per model. 
The snippet below illustrates the text-matching aggregation used in the benchmark.



\end{document}